\documentclass{article}
\usepackage{iclr2027_conference,times}

\usepackage{hyperref}
\usepackage{url}
\usepackage{fontawesome5}
\usepackage{booktabs}
\usepackage{amsmath}
\usepackage{amssymb}
\usepackage{xcolor}
\usepackage{soul}
\usepackage{graphicx}
\usepackage{multirow}
\usepackage{colortbl}
\usepackage{wrapfig}
\usepackage{subcaption}

\newcommand{\mcpicon}[1]{%
  \raisebox{-0.2\height}{#1}\,%
}

\usepackage{enumitem}
\usepackage{listings}
\usepackage[most]{tcolorbox}
\definecolor{promptframe}{HTML}{0F4FA8}
\definecolor{prompttitlebg}{HTML}{DCE5EF}
\definecolor{promptbg}{HTML}{F8FBFF}
\definecolor{codebg}{HTML}{EAF0F6}
\definecolor{codekeyword}{HTML}{AF00DB}
\definecolor{codestring}{HTML}{A31515}
\definecolor{codecomment}{HTML}{008000}
\definecolor{codeident}{HTML}{001080}
\newtcolorbox{promptbox}[1]{breakable, enhanced, colback=promptbg, colframe=promptframe, boxrule=0.6pt, arc=2pt,
  left=6pt, right=6pt, top=4pt, bottom=4pt, title={#1}, fonttitle=\bfseries\small, coltitle=promptframe,
  colbacktitle=prompttitlebg, parbox=false}
\newcommand{\promptsection}[1]{\par\medskip\noindent\textbf{#1}\par\smallskip}
\newcommand{\promptsubsection}[1]{\par\smallskip\noindent\textbf{\textit{#1}}\par\smallskip}
\newcommand{\method}{ProgAgent}

\title{\method: Learning Tool Orchestration through Programmatic Training}

\author{Hao Li$^{1}$\thanks{Equal contribution. Work done during Hao Li's RA appointment at HKUST. \href{mailto:hao612@buaa.edu.cn}{\texttt{hao612@buaa.edu.cn}}.}, Keqing He$^{2}$, Longtao Zheng$^{3}$, Junlong Li$^{1}$, Weihao Zeng$^{1}$, Qian Liu, Yuzhen Huang$^{1,*}$\thanks{Corresponding author: \href{mailto:yhuanghj@connect.ust.hk}{\texttt{yhuanghj@connect.ust.hk}}.} \\
$^{1}$Hong Kong University of Science and Technology\\$^{2}$Beijing University of Posts and Telecommunications \\ $^{3}$Nanyang Technological University  \\
}

\begin{document}

\maketitle
\fancyhead{}
\fancyhead[L]{\small \method: Learning Tool Orchestration through Programmatic Training}
\begin{abstract}
Large language models (LLMs) are increasingly expected to orchestrate diverse tools over long-horizon tasks.
Existing agents typically use Direct Tool Calling (DTC), where the model invokes tools sequentially, and tool responses accumulate in the model context, which can lead to context rot and orchestration breakdown.
In contrast, Programmatic Tool Calling (PTC) invokes tools through executable code and stores intermediate results in runtime variables rather than appending them to the context.
Prior work has explored PTC mainly as an inference-time optimization, but its benefits for agentic training remain unclear, and suitable verifiable training tasks are scarce.
In this work, we introduce \method{}, a full PTC training stack, comprising a pipeline for synthesizing verifiable long-horizon tasks, a lightweight PTC runtime, and a two-stage training recipe.
Using only 215 tasks instantiated at four environment scales, \method{} yields higher pass rates than the evaluated baselines of comparable size across four benchmarks.
On LOCA-bench, \method{}-14B achieves a 30.7\% pass rate at a 64k-token environment state, where these baselines score zero.
Cross-harness experiments show that PTC-trained models outperform DTC-trained models under both evaluation interfaces, suggesting that the benefits of PTC training extend beyond the interface. 
More details are on our \href{https://llleoli.github.io/ProgAgent/}{\faGlobe\ project page}.

\end{abstract}

\section{Introduction}
\label{sec:intro}

Large language models (LLMs) increasingly excel in deep research~\citep{openai2025deepresearch, google2025geminideepresearch} and software engineering~\citep{jimenez2024swebench, lin_scaling_agents_2026}.
However, they struggle in long-horizon open-world tool environments~\citep{li2025toolathlon}.
Agents relying on \emph{Direct Tool Calling} (DTC) invoke tools across multiple turns, maintaining execution logic and tool results in context.
As raw tool outputs accumulate in the model context, they can contribute to context rot~\citep{lee_context_rot_2025, chroma_context_rot_2025} and orchestration breakdown~\citep{zeng2026loca}. 

Complex tool orchestration often requires data transformation, conditional branching, and error handling.
While difficult to express through sequential tool calls, these structures are native to programming languages.
\emph{Programmatic Tool Calling} (PTC) provides an interface for applying models' coding abilities to tool orchestration~\citep{anthropic_advanced_tool_use_2025}.
Under PTC, the model generates code that invokes tools and processes intermediate results within a sandbox, storing tool returns in variables and exposing explicitly printed outputs to the model context, as shown in Figure~\ref{fig:transfer}.

While PTC has mainly been explored as an inference-time optimization~\citep{anthropic_advanced_tool_use_2025, karten2026prime}, we argue that its advantages make it an effective training paradigm.
By representing tool orchestration as code, PTC may enable foundation models to transfer their pre-trained coding capabilities to complex tool-use tasks.
PTC can also limit context growth by retaining intermediate results in runtime variables and selectively printing outputs.
We discuss these advantages in \S\ref{sec:ptc_alignment}.

Despite this potential, three gaps limit systematic PTC training for long-horizon tasks.
\emph{First}, publicly available tasks designed for PTC training remain scarce. Training agents for programmatic orchestration requires long-horizon tasks involving diverse tools and verifiable rewards.
\emph{Second}, open-source harnesses~\citep{shi2026programming,karten2026prime} embed PTC within general-purpose agent systems.
Controlled studies of PTC training require a runtime that preserves execution state across turns and isolates episodes during RL rollouts.
\emph{Third}, the benefits of PTC as a training paradigm remain unclear. Existing inference-time evaluations~\citep{anthropic_advanced_tool_use_2025} do not establish whether training agents to use PTC improves tool orchestration or whether the capabilities learned under PTC transfer to DTC.

\begin{figure}[t]
\centering
\includegraphics[width=\textwidth]{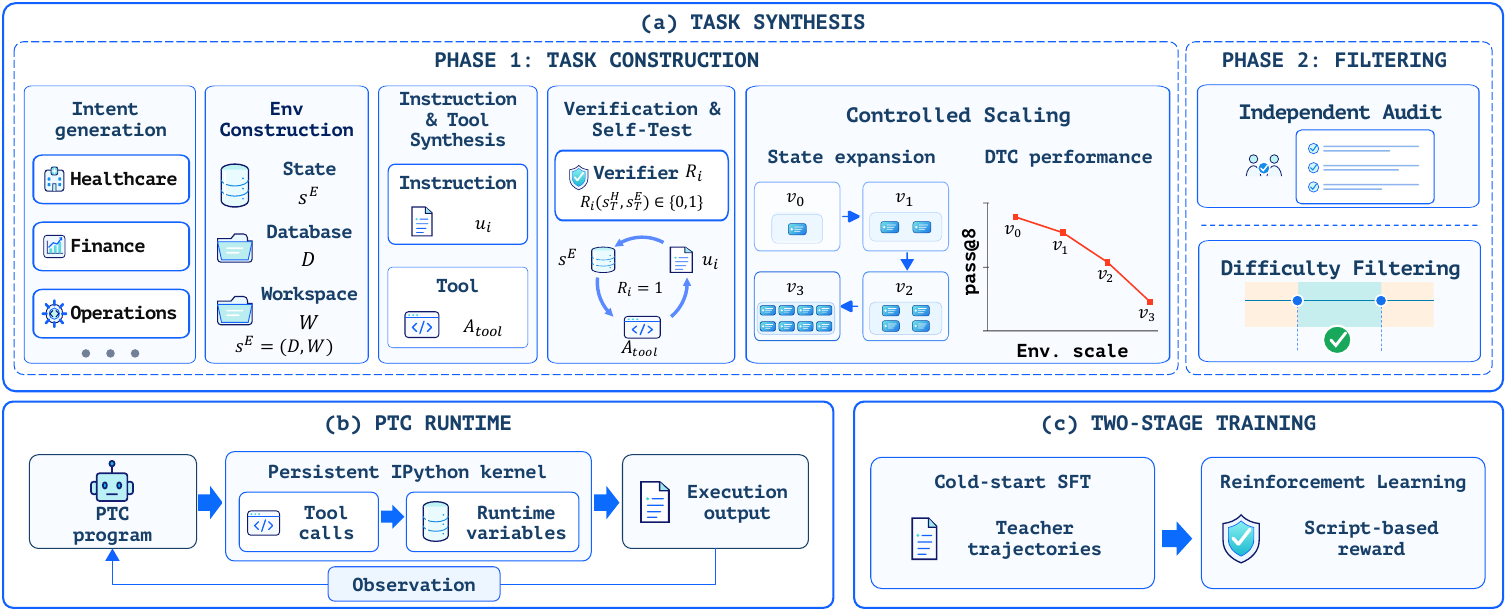}
\caption{\textbf{Overview of \method.}
\textbf{(a) Task synthesis.} We construct verifiable tasks, scale their environment states from $v_0$ to $v_3$ while keeping instructions, task logic, and tool schemas fixed, and apply independent auditing and difficulty filtering.
The scaling plot shows DeepSeek V4 Pro's declining DTC pass@8 as environment size increases.
\textbf{(b) PTC runtime.} A persistent IPython kernel executes tool-calling programs, retains intermediate results in variables across turns, and returns printed outputs to the agent.
\textbf{(c) Two-stage training.} Cold-start SFT on teacher trajectories is followed by RL with binary rewards from Python task verifiers.
}
\label{fig:task_synthesis_pipeline}
\end{figure}

In this paper, we introduce \method{}, a training stack for studying PTC as a training paradigm for long-horizon tool orchestration (Figure~\ref{fig:task_synthesis_pipeline}).
We synthesize verifiable long-horizon tasks, scale their environment states to control task difficulty, and pair each task with a Python verifier that returns a binary reward (\S\ref{sec:task_synthesis}).
To support reinforcement learning (RL) in these environments, we build a lightweight PTC runtime that preserves execution state across turns and isolates episodes during rollouts (\S\ref{sec:runtime}).
Using these environments and the runtime, we train models through cold-start supervised fine-tuning (SFT) followed by RL (\S\ref{sec:training}).

We evaluate \method{} on four benchmarks using 8B and 14B models trained on 215 tasks instantiated at four environment scales (\S\ref{sec:main_results}).
Although the base models perform poorly under PTC, training with \method{} improves performance across all four benchmarks.
On MCPMark~\citep{wu2025mcpmark} and Toolathlon-Verified~\citep{li2025toolathlon}, our 8B and 14B models outperform the evaluated baselines of comparable size.
On LOCA-bench~\citep{zeng2026loca}, \method{}-14B achieves 30.7\% at an environment-state size of 64k tokens, where the comparable-size baselines score 0\%.
On MCP-Atlas~\citep{bandi2026mcpatlas}, \method{}-14B performs comparably to Claude Sonnet-4.5 evaluated under DTC.
PTC achieves higher pass rates than Hybrid (PTC+DTC) training, suggesting that a uniform programmatic action space may benefit tool orchestration. 
Cross-harness evaluations also show that PTC-trained models outperform DTC-trained models under both interfaces, suggesting that the benefits of PTC training extend to direct tool calling (\S\ref{sec:what_transfers}).

\section{Reframing Tool Use as Code}

\subsection{Multi-Turn Agentic Interaction as a Partially Observable Markov Decision Process}
\label{sec:pomdp}
We formulate multi-turn agentic interaction with external environments as a Partially Observable Markov Decision Process (POMDP)~\citep{aastrom1965optimal}, following AgentSkiller~\citep{sun2026agentskiller}. A POMDP is defined by a tuple
\begin{equation*}
    \mathcal{M} = (\mathcal{S}, \mathcal{A}, \mathcal{O}, P, R),
\end{equation*}
where $\mathcal{S}$, $\mathcal{A}$, and $\mathcal{O}$ are the state, action, and observation spaces, respectively, and $P$ is the transition function.
In our setting, $R$ is a programmatic verifier that returns a binary reward at the end of an episode. 
At each turn $t$, the agent receives an observation $o_t \in \mathcal{O}$ rather than the full state $s_t \in \mathcal{S}$ and takes an action $a_t \in \mathcal{A}$.
The environment  transitions to $s_{t+1}$ and returns an observation $o_{t+1}$. In our setting, the state decomposes into an environment component and an interaction component:
\[
    s_t = (s_t^E, s_t^H),
\]
where the environment state $s_t^E$ contains databases, workspace files, and executable tools, and $s_t^H$ tracks the agent's interaction history, including task instructions, returned observations, and reasoning context. The action space partitions into tool actions and natural-language responses:
\[
    \mathcal{A} = \mathcal{A}_{\mathrm{tool}} \cup \mathcal{A}_{\mathrm{resp}}.
\]
Tool actions may modify $s_t^E$. Agent actions and observations are appended to the interaction history $s_t^H$. At the end of an episode of $T$ turns, the verifier returns a reward $R(s_T^H, s_T^E) \in \{0, 1\}$.

\subsection{From Direct Tool Calling to Programmatic Tool Calling}
\label{sec:dtc_vs_ptc}
\begin{figure}[t]
\centering
\includegraphics[width=\textwidth]{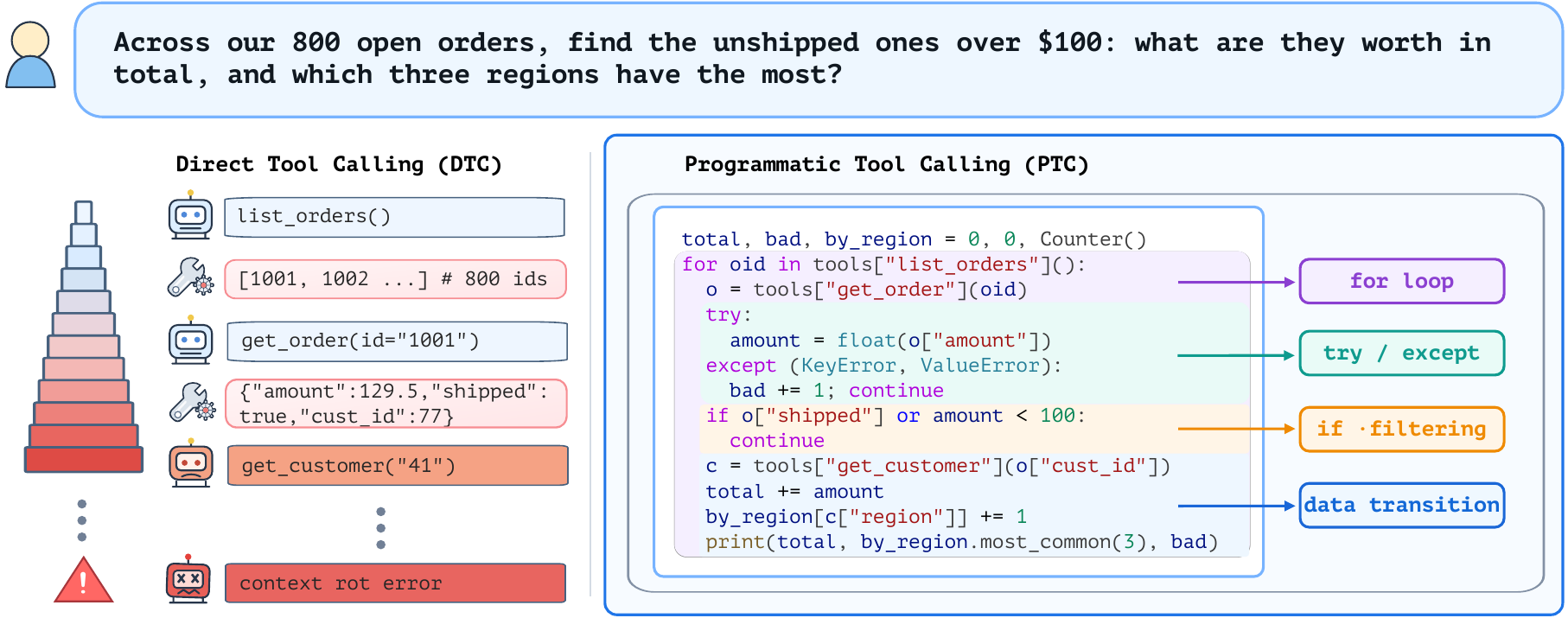}
\caption{Direct tool calling (DTC) versus programmatic tool calling (PTC). DTC accumulates raw observations in the context, whereas PTC orchestrates tools via code and selectively prints outputs.}
\label{fig:transfer}
\end{figure}
Within this framework, we distinguish between two harnesses: direct tool calling (DTC) and programmatic tool calling (PTC) (Figure~\ref{fig:transfer}).
\paragraph{Direct tool calling (DTC).} As shown in  Figure~\ref{fig:transfer} (left), under DTC, the policy $\pi$ generates an action $a_t \in \mathcal{A}_{\mathrm{tool}}$ at each turn $t$, and the environment transitions as $s_t^E \xrightarrow{a_t} s_{t+1}^E$.
The action $a_t$ and its returned observation $o_{t+1}$ are appended to the interaction context: $s_{t+1}^H = s_t^H \oplus a_t \oplus o_{t+1}$, where $\oplus$ denotes concatenation.
Task control flow is expressed through a sequence of tool calls, with execution logic and tool results maintained in context.
The context grows as $|s_T^H| = |s_0^H| + \sum_{t=0}^{T-1} (|a_t| + |o_{t+1}|)$.
As raw tool outputs accumulate in context, they can contribute to context rot~\citep{lee_context_rot_2025, chroma_context_rot_2025} and orchestration breakdown~\citep{li2025toolathlon}.

\paragraph{Programmatic tool calling (PTC).}  As shown in  Figure~\ref{fig:transfer} (right), under PTC, the policy $\pi$ generates a program $\rho_t \sim \pi(\cdot \mid s_t^H)$ that runs inside a sandbox.
The program $\rho_t$ invokes tools through calls $a_{t,1}, a_{t,2}, \ldots$ and can include control flow, data transformation, and exception handling.
Its execution induces the environment transition $s_t^E \xrightarrow{\rho_t} s_{t+1}^E$.
Intermediate tool results are stored in runtime variables and retained across turns.
Only the program and its printed output $r_{\rho_t}$ are appended to the interaction context: $s_{t+1}^H = s_t^H \oplus \rho_t \oplus r_{\rho_t}$.
By selectively printing outputs, PTC can limit the number of tool-output tokens entering the context (\S\ref{sec:training_analysis}).

\subsection{PTC as an Effective Training Paradigm}
\label{sec:ptc_alignment}
Beyond inference efficiency, PTC offers two advantages that may make it an effective training paradigm.
First, PTC may enable models to transfer their pre-trained coding capabilities to agentic tasks.
During pre-training, foundation models learn to use control flow, data transformation, and error handling in code, which are also useful for tool orchestration.
With environment tools exposed as callable functions, the generated program $\rho_t$ can use these structures to coordinate multiple interrelated tool calls within a single turn.
Second, PTC provides a mechanism for context management.
By retaining intermediate results in runtime variables and selectively printing outputs $r_{\rho_t}$ to the interaction context $s_t^H$, PTC can limit context growth.
In this work, we explore whether this paradigm enables models to achieve better agentic performance with fewer training tasks.
We also investigate whether models can learn this context filtering behavior during training to mitigate context rot.
\section{\method: From Task Synthesis to Agentic Training}

We build \method, a training stack with three components:
a task synthesis pipeline for producing PTC-tailored, verifiable long-horizon tasks with controllable difficulty (\S\ref{sec:task_synthesis});
a lightweight PTC runtime that uses a persistent IPython kernel to preserve execution state across turns and Bubblewrap~\citep{bubblewrap2025} for sandbox isolation (\S\ref{sec:runtime});
and a training recipe consisting of cold-start SFT followed by RL to improve long-horizon tool orchestration (\S\ref{sec:training}).
\subsection{Synthesizing Verifiable Long-Horizon Agentic Tasks}
\label{sec:task_synthesis}

\subsubsection{Design Principles}
\label{sec:design_principles}
\paragraph{Long-Horizon Environment State Scaling.}
Existing environment synthesis pipelines~\citep{song2026envscaler,dong2026agent} primarily make tasks harder by increasing tool diversity or orchestration complexity.
However, these changes do not provide direct control over environment-state size and task difficulty, limiting their ability to optimize model capabilities and behavior under context rot.
Inspired by the concept of description length~\citep{hutter2000theory,legg2007universal,zeng2026loca}, we hold the task instruction, underlying logic, and tool schemas constant while scaling the environment state at each discrete level $j$ ($|s^{E,(j+1)}| \approx 2 \cdot |s^{E,(j)}|$).
As $s^E$ grows, the agent must retrieve and process information from a larger environment.
Under DTC, the resulting tool observations are appended directly to the interaction context $s^H$.
As the environment scales from $v_0$ to $v_3$, DeepSeek V4 Pro's~\citep{deepseekai2026deepseekv4} pass@8 on our tasks drops from 0.95 to 0.42 under DTC (Figure~\ref{fig:task_synthesis_pipeline}).
Under PTC, its pass@8 remains above 0.90 across all four scales (Figure~\ref{fig:env_scaling_harness}).

\paragraph{Programmatic Task Design.}
We design tasks with tool orchestration patterns that can be naturally expressed through programs.
Specifically, we distribute environment data across related tables in a relational database and place task constraints in unstructured files.
Solving these tasks involves interpreting these constraints and combining information from several sources.
Loops and conditional branching can help coordinate tool calls during multi-hop aggregation.

\paragraph{Rigorous Task Verifiability.}
We pair each synthesized task $i$ with a programmatic verifier $R_i(s_T^H, s_T^E) \in \{0, 1\}$.
Implemented as a Python script, the verifier checks the final response and final environment state against the task's evaluation criteria, returning 1 if and only if all criteria are satisfied.
We verify task solvability by requiring an oracle solution to receive full reward using only the provided tools.
If the oracle fails verification, an automated repair loop revises the task environment, tools, instruction, or verifier until the task passes this self-test.

\subsubsection{Task Construction Pipeline}
\label{sec:construction_pipeline}

Each task is built in five stages (Phase~1 of Figure~\ref{fig:task_synthesis_pipeline}).
First, an LLM generates user intents from 23 curated domains detailed in Appendix~\ref{sec:appendix_intents}.
A constructor agent then builds a relational database $D$ and a file workspace $W$ using scripts with a scaling configuration that controls environment-state size.
Next, this agent synthesizes the action space $\mathcal{A}_{\mathrm{tool}}$ as a set of single-purpose tools whose Python implementations are hidden from the task-solving agent, together with an instruction $u_i$ that specifies the task goals and required outputs.
It then generates a verifier $R_i$ and an oracle solution, iteratively repairing the task until the oracle solution receives full reward.
Finally, the agent adjusts the scaling configuration to generate four variants $\mathcal{M}_i^{(0)},\dots,\mathcal{M}_i^{(3)}$ (denoted $v_0$--$v_3$), increasing the environment-state size $|s^{E,(j)}|$ at each level while keeping the task instruction, underlying logic, and tool schemas fixed.
We provide more details in Appendix~\ref{sec:appendix_pipeline_details}.

\subsubsection{Verification-based Filtering}
\label{sec:filtering}

After oracle verification, we audit each task for consistency and filter tasks by difficulty (Phase~2 of Figure~\ref{fig:task_synthesis_pipeline}; details in Appendix~\ref{sec:appendix_filtering_details}).
First, an independent harness agent checks that the tools, environment state, and verifier are consistent with the task instruction.
Tasks with identified issues are repaired and audited again.
Next, for each task and each scale, we sample $K=8$ trajectories from DeepSeek V4 Pro under the PTC harness.
We retain a task only if its pass rate lies in $(0,0.8)$ at every scale, selecting tasks that remain challenging but solvable.
Finally, each surviving task $\mathcal{M}_i$ is packaged into three core artifacts: a data generator, an environment module, and an instruction $u_i$.

\subsection{A Lightweight PTC Runtime}
\label{sec:runtime}
Following LOCA-bench~\citep{zeng2026loca}, we expose the PTC runtime to the model as an MCP server.
At the start of each episode, we run the data generator at a target scale $j$ to initialize the environment state $s_0^E = (D^{(j)}, W^{(j)})$ and provide the model with the instruction $u_i$.
The runtime uses a persistent, sandboxed IPython kernel.
A tool proxy routes calls from each generated program $\rho_t$ to the corresponding environment tools in $\mathcal{A}_{\mathrm{tool}}$ and returns their results as native Python objects.
The kernel executes $\rho_t$, returns its printed output as $r_{\rho_t}$, and preserves runtime variables across turns.
The sandbox prevents the agent from accessing the verifier implementation and oracle solution (Appendix~\ref{sec:appendix_runtime}).
At the end of each episode, the verifier $R_i(s_T^H, s_T^E)$ evaluates the final agent response and environment state to produce a binary reward.

\subsection{Training with \method}
\label{sec:training}

After filtering, the synthesis pipeline retains 215 tasks, each instantiated at the four environment scales ($v_0$--$v_3$) defined in \S\ref{sec:construction_pipeline}.
Training proceeds in two stages: cold-start SFT on teacher trajectories, followed by RL via GSPO~\citep{zheng2025stabilizing,gspo}.
At the end of each rollout, the verifier $R_i(s_T^H, s_T^E)$ evaluates the final response and environment state to return a binary reward.
We apply this training recipe across three runtime harnesses:
(1) \textbf{DTC}: exposes environment tools through standard schemas for direct calls;
(2) \textbf{PTC}: requires all environment tool calls to go through the Python sandbox;
(3) \textbf{Hybrid}: exposes both interfaces, allowing the model to interleave direct calls and program executions.
To isolate the effect of the harness, we use the same task pool and training hyperparameters across all three settings (details in \S\ref{sec:exeperiment_setup} and Appendix~\ref{sec:appendix_hyperparams}).

\section{Experiments}
\label{sec:experiments}

\subsection{Experimental Setup}
\label{sec:exeperiment_setup}
\paragraph{Synthesis and Data Collection.}
We generate task intents using Gemini~3.0~Flash~\citep{google2025gemini3}, and the constructor and harness agents rely on Claude Opus~4.6~\citep{anthropic2026opus46} via Claude Code.
For environment difficulty filtering, we sample 8 DeepSeek V4 Pro~\citep{deepseekai2026deepseekv4} trajectories per candidate task at each scale ($v_0$--$v_3$) under PTC, retaining only tasks whose pass rate lies in $(0,0.8)$ at every scale.
The resulting 215 tasks are each instantiated at 4 scales, yielding a shared training pool of 860 task instances.
Appendix~\ref{sec:appendix_intents} shows the details of task distribution.

\paragraph{Training Settings \& Details.}
We train Qwen3~\citep{qwen3technicalreport} (8B and 14B) across the DTC, PTC, and Hybrid settings and evaluate each model using its training harness.
Models are first fine-tuned on the cold-start datasets for 3 epochs (batch size 64), then trained with GSPO~\citep{gspo} for 150 steps, sampling 16 rollouts per task for 32 tasks at each step (full details in Appendix~\ref{sec:appendix_hyperparams}).

\paragraph{Benchmarks \& Baselines.}
We evaluate our models across three agentic benchmarks: MCPMark~\citep{wu2025mcpmark}, MCP-Atlas~\citep{bandi2026mcpatlas}, and Toolathlon-Verified~\citep{li2025toolathlon}.
To stress-test robustness under extreme environment context growth, we further use LOCA-bench~\citep{zeng2026loca} at 8k--64k environment-state sizes, a benchmark for evaluating long-horizon agents in realistic settings.
Each reported pass rate averages four independent runs. Our baselines include frontier proprietary models~\citep{openai2025gpt52,anthropic2025sonnet45,google2025gemini3}, open-source foundation models~\citep{qwen3technicalreport,deepseekai2025deepseekv32}, and six trained baselines summarized in Table~\ref{tab:baseline_overview}.
In Table~\ref{tab:main_results}, we evaluate baselines under the DTC harness. Additional evaluations of selected baselines under the PTC harness are reported in Table~\ref{tab:ablations_loca}.

\begin{table}
  \centering
  \caption{\textbf{Training-data scales across baselines} (traj.: trajectories).
  Simulator aggregates synthesis settings; EnvScaler counts SFT + RL samples~\citep{xu2026envfactory}.}
  \label{tab:baseline_overview}
  \small
  \setlength{\tabcolsep}{2pt}
  \renewcommand{\arraystretch}{1.1}
  \begin{tabular}{@{}p{0.07\textwidth}*{7}{>{\centering\arraybackslash}p{\dimexpr0.93\textwidth/7-4pt\relax}}@{}}
    \toprule
    \textbf{Method} & TOUCAN & Simulator & EnvScaler & AWM & Agent-World & EnvFactory & \textbf{\method} \\
    & {\footnotesize\citep{xu2025toucan}}
    & {\footnotesize\citep{li2025simulating}}
    & {\footnotesize\citep{song2026envscaler}}
    & {\footnotesize\citep{wang2026awm}}
    & {\footnotesize\citep{dong2026agent}}
    & {\footnotesize\citep{xu2026envfactory}}
    & {\footnotesize (ours)} \\
    \midrule
    \textbf{Tasks} & 1.53M traj. & 135k traj. & 11,572 & 3,315 & 2,000+ & 2,575 & $215 \times 4$ \\
    \bottomrule
  \end{tabular}
\end{table}

\subsection{Main Results}
\label{sec:main_results}
\begin{table}[!t]
  \centering
  \caption{
      \textbf{Main results.}
      Mean pass rates (\%) on four benchmarks.
      Best and second-best scores within each 8B/14B comparison are bolded and underlined.
      Evaluation protocols and sources of reported scores are detailed in \S\ref{sec:exeperiment_setup} and Appendix~\ref{sec:appendix_eval_protocol}.
  }
  \label{tab:main_results}
  \setlength\tabcolsep{2.5pt}
  \renewcommand{\arraystretch}{1.2}
  \resizebox{\textwidth}{!}{%
  \begin{tabular}{>{\raggedright\arraybackslash}p{2.9cm}
    ccccc !{\hspace{4pt}}
    *{5}{>{\centering\arraybackslash}p{0.9cm}} !{\hspace{4pt}}
    >{\centering\arraybackslash}p{1.8cm} !{\hspace{4pt}}
    >{\centering\arraybackslash}p{1.8cm}}
  \toprule
  \multirow{2}[2]{*}{\textbf{Model}} &
  \multicolumn{5}{c}{\textbf{MCPMark}} &
  \multicolumn{5}{c}{\textbf{LOCA-bench}} &
  \multirow{2}[2]{*}{\textbf{Toolathlon}} &
  \multirow{2}[2]{*}{\textbf{MCP-Atlas}} \\
  \cmidrule(lr){2-6} \cmidrule(lr){7-11}
   & \mcpicon{\includegraphics[height=1.2em]{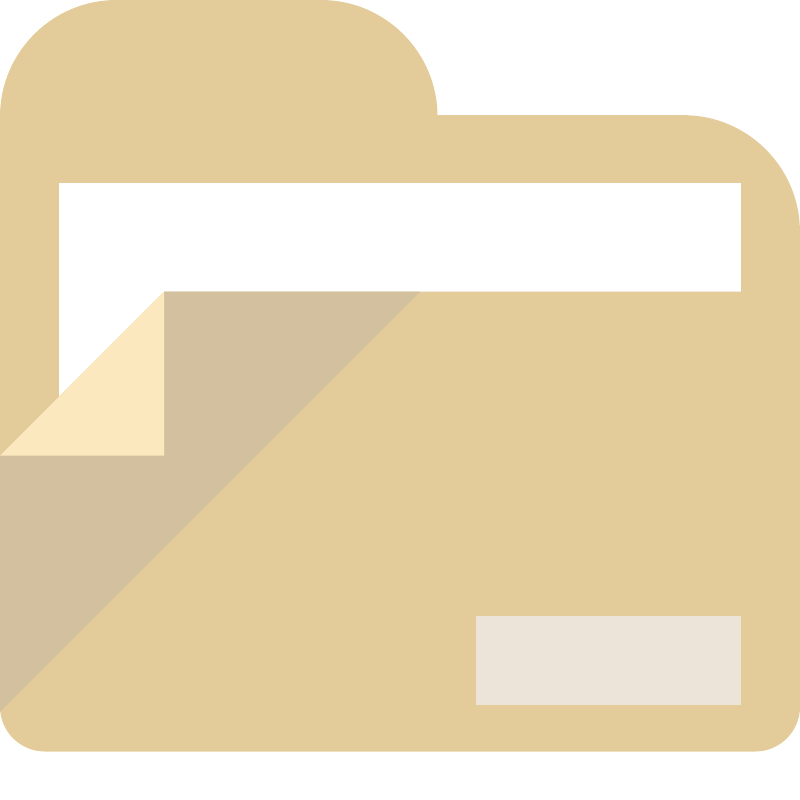}}File. & \mcpicon{\includegraphics[height=1.2em]{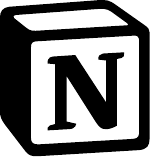}}Notion & \mcpicon{\includegraphics[height=1.2em]{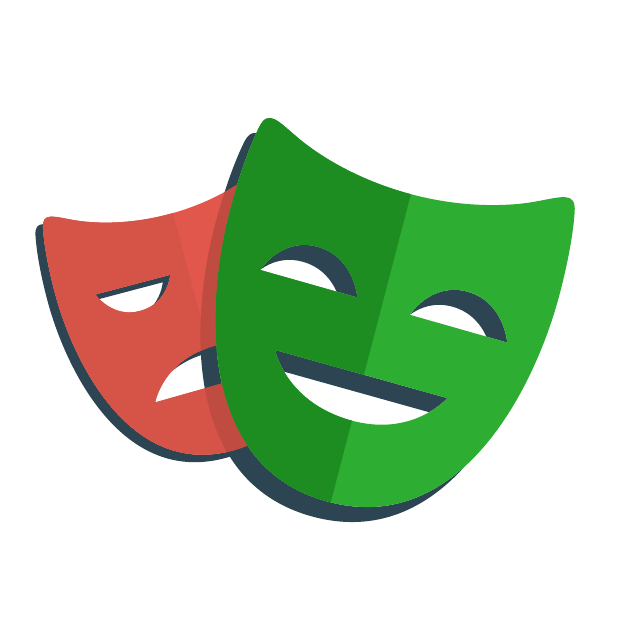}}Play. & \mcpicon{\includegraphics[height=1.2em]{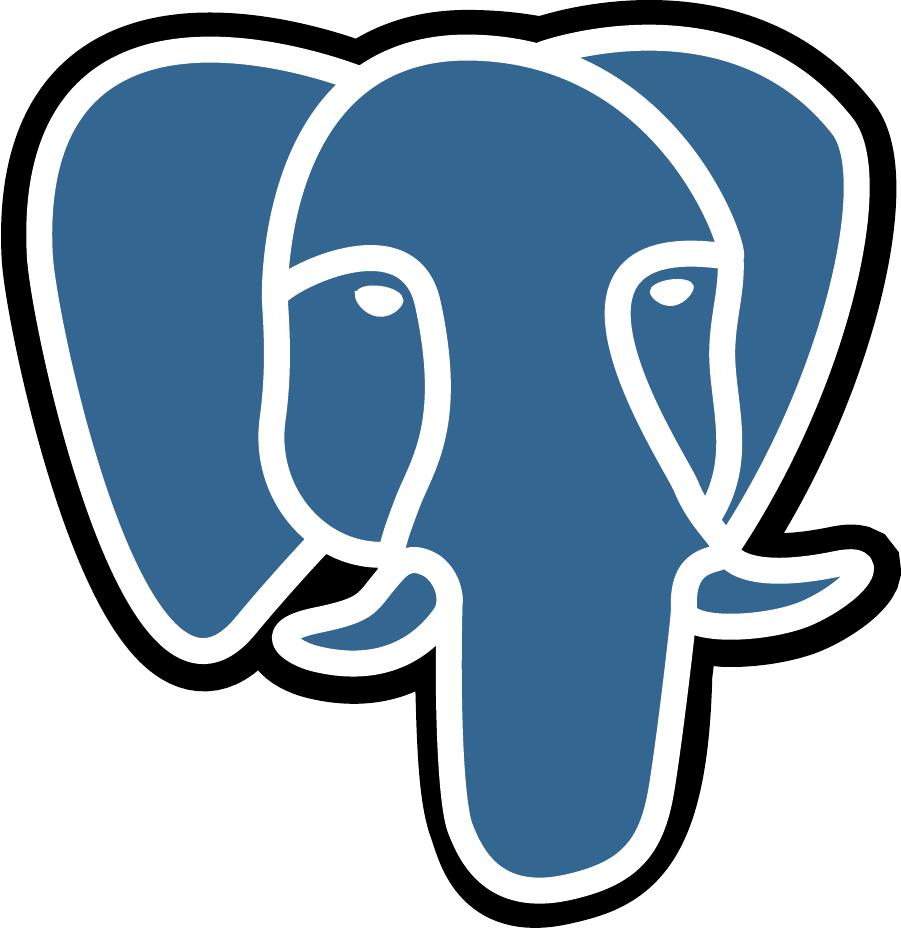}}Post. & \textbf{Avg.}
   & 8k & 16k & 32k & 64k & \textbf{Avg.}
   & & \\
  \midrule
  \rowcolor[HTML]{eaf3fb}
  \multicolumn{13}{c}{\textbf{\textit{Frontier Proprietary Models}}} \\
  GPT-5.2
    & 60.0 & 42.9 & 40.0 & 66.7 & 52.4 & 72.0 & 70.7 & 60.0 & 52.0 & 63.7 & 41.7$^{\dagger}$ & 67.6 \\
  Claude Sonnet-4.5
    & 32.5 & 25.0 & 27.0 & 50.0 & 33.6 & 58.7 & 60.0 & 36.0 & 26.7 & 45.3 & 38.9$^{\dagger}$ & 43.8 \\
  Gemini-3 Pro
    & 56.7 & 43.8 & 40.0 & 70.2 & 52.7 & 26.7 & 22.7 & 17.3 & 9.3 & 19.0 & 36.4$^{\dagger}$ & 54.1 \\
  \midrule
  \rowcolor[HTML]{eaf3fb}
  \multicolumn{13}{c}{\textbf{\textit{Open-Source Foundation Models}}} \\
  Qwen3-8B
    & 3.3 & 0.0 & 4.0 & 4.8 & 3.0 & 1.3 & 0.0 & 0.0 & 0.0 & 0.3 & 2.8 & 7.6 \\
  Qwen3-14B
    & 3.3 & 0.0 & 0.0 & 9.5 & 3.2 & 1.3 & 0.0 & 0.0 & 0.0 & 0.3 & 1.9 & 12.4 \\
  Qwen3-235B-A22B
    & 13.3 & 10.7 & 0.0 & 4.8 & 7.2 & 2.7 & 1.3 & 0.0 & 0.0 & 1.0 & 3.7 & 12.0 \\
  DeepSeek-V3.2
    & 36.7 & 45.5 & 17.0 & 66.6 & 41.5 & 78.7 & 80.0 & 61.3 & 45.3 & 66.3 & 35.2$^{\dagger}$ & 56.2 \\
  \midrule
  \rowcolor[HTML]{eaf3fb}
  \multicolumn{13}{c}{\textbf{\textit{Open-Source Baselines}}} \\
  TOUCAN-7B
    & 0.0 & 0.0 & 0.0 & 4.8 & 1.2 & 0.0 & 0.0 & 0.0 & 0.0 & 0.0 & 0.0 & 7.0 \\
  Simulator-8B
    & 3.3 & 0.0 & 4.0 & 4.8 & 3.0 & 0.0 & 0.0 & 0.0 & 0.0 & 0.0 & 0.0 & 0.8 \\
  AWM-8B
    & 3.3 & 0.0 & 4.0 & 4.8 & 3.0 & 0.0 & 0.0 & 0.0 & 0.0 & 0.0 & 1.9 & 8.0 \\
  EnvFactory-8B
    & 10.0 & 0.0 & 4.0 & 9.5 & 5.9 & \underline{2.7} & \underline{1.3} & 0.0 & 0.0 & \underline{1.0} & 1.9 & 11.8 \\
  EnvScaler-8B
    & 10.0 & 0.0 & 4.0 & 9.5 & 5.9 & 1.3 & 0.0 & 0.0 & 0.0 & 0.3 & \underline{2.8} & \underline{13.0} \\
  Agent-World-8B
    & \underline{13.3} & \underline{3.6} & 4.0 & \underline{19.1} & \underline{10.0} & -- & -- & -- & -- & -- & -- & -- \\
  \addlinespace[2pt]
  AWM-14B
    & 3.3 & 0.0 & 4.0 & 9.5 & 4.2 & \underline{5.3} & \underline{2.7} & 0.0 & 0.0 & \underline{2.0} & \underline{1.9} & \underline{11.0} \\
  Agent-World-14B
    & \underline{16.6} & \underline{3.6} & 4.0 & \underline{38.1} & \underline{15.6} & -- & -- & -- & -- & -- & -- & -- \\
  \midrule
  \rowcolor[HTML]{eaf3fb}
  \multicolumn{13}{c}{\textbf{\textit{Our Method}}} \\
  \textbf{\method-8B}
    & \textbf{33.3} & \textbf{14.3} & 4.0 & \textbf{21.4} & \textbf{18.3} & \textbf{34.7} & \textbf{24.0} & \textbf{21.3} & \textbf{17.3} & \textbf{24.3} & \textbf{9.3} & \textbf{38.6} \\
  \textbf{\method-14B}
    & \textbf{46.7} & \textbf{25.0} & 4.0 & \textbf{42.9} & \textbf{29.7} & \textbf{41.3} & \textbf{40.0} & \textbf{41.3} & \textbf{30.7} & \textbf{38.3} & \textbf{17.1} & \textbf{45.6} \\
  \bottomrule
  \end{tabular}
  }
\end{table}

As Table~\ref{tab:main_results} shows, \method{} outperforms comparable baselines across diverse agentic benchmarks.
On MCPMark, \method{} achieves average pass rates of 18.3\% at 8B and 29.7\% at 14B, respectively.
On Toolathlon-Verified, \method{} reaches 9.3\% at 8B and 17.1\% at 14B, compared with at most 2.8\% and 1.9\% among the 8B and 14B baselines.
Moreover, on MCP-Atlas, \method-14B achieves 45.6\%, comparable to Claude Sonnet-4.5 (43.8\%) under the DTC harness.
The gap is most pronounced on LOCA-bench: the evaluated comparable-size baselines score zero at 32k and 64k, while \method{} mitigates context rot to maintain pass rates between 17.3\% and 30.7\% at 64k.
\begin{table}[!t]
  \centering
  \caption{
      \textbf{Training paradigms across benchmarks.}
      Final pass rates (\%) with the corresponding base model's scores under the same harness shown in parentheses.
      Each trained model is evaluated under its training harness.
      The best score for each benchmark within each backbone is bolded.
  }
  \label{tab:paradigm_comparison}
  \setlength\tabcolsep{3pt}
  \renewcommand{\arraystretch}{1.15}
  \resizebox{\textwidth}{!}{%
\begin{tabular}{l l ccccc ccc}
\toprule
\multirow{2}[2]{*}{\textbf{Model}} &
\multirow{2}[2]{*}{\textbf{Paradigm}} &
\multicolumn{5}{c}{\textbf{MCPMark}} &
\multirow{2}[2]{*}{\textbf{LOCA-bench}} &
\multirow{2}[2]{*}{\textbf{Toolathlon}} &
\multirow{2}[2]{*}{\textbf{MCP-Atlas}} \\
\cmidrule(lr){3-7}
 & & \mcpicon{\includegraphics[height=1.2em]{figures/icons/filesystem.pdf}}File. & \mcpicon{\includegraphics[height=1.2em]{figures/icons/notion.pdf}}Notion & \mcpicon{\includegraphics[height=1.2em]{figures/icons/playwright.pdf}}Play. & \mcpicon{\includegraphics[height=1.2em]{figures/icons/postgres.pdf}}Post. & \textbf{Avg.} & & & \\
\midrule
\multirow{3}{*}{8B} & DTC & 6.7~{\scriptsize (3.3)} & 3.6~{\scriptsize (0.0)} & 4.0~{\scriptsize (4.0)} & 9.5~{\scriptsize (4.8)} & 6.0~{\scriptsize (3.0)} & 4.0~{\scriptsize (0.3)} & 2.8~{\scriptsize (2.8)} & 19.0~{\scriptsize (7.6)} \\
 & Hybrid & 30.0~{\scriptsize (13.3)} & 5.4~{\scriptsize (0.0)} & 4.0~{\scriptsize (0.0)} & 19.1~{\scriptsize (4.8)} & 14.6~{\scriptsize (4.5)} & 21.7~{\scriptsize (0.3)} & 7.4~{\scriptsize (0.9)} & 30.6~{\scriptsize (8.4)} \\
 & PTC & 33.3~{\scriptsize (5.0)} & 14.3~{\scriptsize (0.0)} & 4.0~{\scriptsize (0.0)} & 21.4~{\scriptsize (4.8)} & \textbf{18.3}~{\scriptsize (2.5)} & \textbf{24.3}~{\scriptsize (0.0)} & \textbf{9.3}~{\scriptsize (0.0)} & \textbf{38.6}~{\scriptsize (3.6)} \\
\cmidrule(l){2-10}
\multirow{3}{*}{14B} & DTC & 10.0~{\scriptsize (3.3)} & 3.6~{\scriptsize (0.0)} & 4.0~{\scriptsize (0.0)} & 14.3~{\scriptsize (9.5)} & 8.0~{\scriptsize (3.2)} & 9.0~{\scriptsize (0.3)} & 5.6~{\scriptsize (1.9)} & 29.7~{\scriptsize (12.4)} \\
 & Hybrid & 43.3~{\scriptsize (3.3)} & 16.1~{\scriptsize (0.0)} & 4.0~{\scriptsize (0.0)} & 38.1~{\scriptsize (9.5)} & 25.4~{\scriptsize (3.2)} & 31.3~{\scriptsize (0.7)} & 14.7~{\scriptsize (3.7)} & 38.6~{\scriptsize (11.4)} \\
 & PTC & 46.7~{\scriptsize (6.7)} & 25.0~{\scriptsize (5.0)} & 4.0~{\scriptsize (0.0)} & 42.9~{\scriptsize (4.8)} & \textbf{29.7}~{\scriptsize (4.1)} & \textbf{38.3}~{\scriptsize (0.0)} & \textbf{17.1}~{\scriptsize (0.9)} & \textbf{45.6}~{\scriptsize (3.8)} \\
\bottomrule
\end{tabular}%
  }
\end{table}

Table~\ref{tab:paradigm_comparison} compares each training recipe with its untrained backbone under the same harness.
For untrained backbones, switching to PTC provides limited benefits across benchmarks and often yields the lowest scores among the three harnesses.
Training reverses this pattern:
after SFT and RL, the PTC recipe achieves the highest aggregate score on each benchmark at both model scales.
\method-14B reaches 29.7\% on MCPMark and 45.6\% on MCP-Atlas, compared with 8.0\% and 29.7\% for DTC, respectively. 
\method-8B also outperforms the 8B DTC model on all four benchmarks.
Cross-harness evaluation on LOCA-bench and MCP-Atlas (Table~\ref{tab:dtc_transfer_loca} in \S\ref{sec:what_transfers}) further demonstrates that the gains extend beyond the inference interface to the training paradigm itself.
\begin{figure}[!t]
\centering
\includegraphics[width=\textwidth]{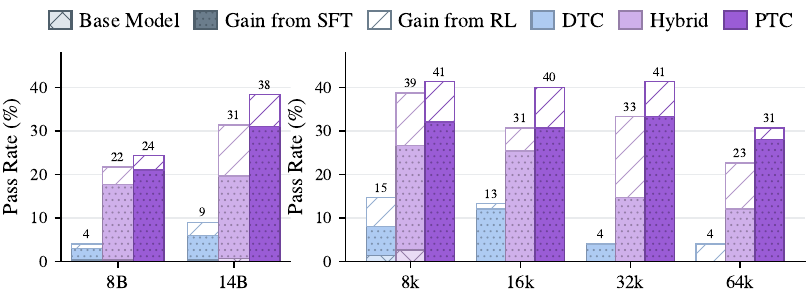}
\caption{\textbf{Training paradigms on LOCA-bench.}
Each bar decomposes the final pass rate into the untrained backbone's score and the gains from SFT and RL.
Labels above the bars show rounded final pass rates; exact values are reported in Table~\ref{tab:ablations_loca}.
\textbf{Left:} Results for the 8B and 14B backbones, averaged over four environment-state sizes.
\textbf{Right:} Results for the 14B backbone at each environment-state size, showing PTC's robustness as the environment grows. 
}
\label{fig:paradigm_ablation}
\end{figure}

Figure~\ref{fig:paradigm_ablation} compares the three training paradigms on LOCA-bench, showing gains from both SFT and RL.
As the environment state scales, DTC suffers from context rot. In contrast, PTC effectively mitigates this degradation and sustains robust performance, achieving average pass rates of 24.3\% (8B) and 38.3\% (14B).
Furthermore, despite providing a more flexible action space during both training and evaluation, the Hybrid paradigm performs worse than PTC.
We hypothesize that PTC provides a unified tool interaction interface which lowers learning difficulty and enables the model to learn a more generalized representation. While a similar design is adopted in DeepSeek Harness \citep{shi2026programming}, we substantiate it with empirical observations.
\section{Analysis}
\label{sec:analysis}
\begin{figure}[!t]
\centering
\includegraphics[width=\textwidth]{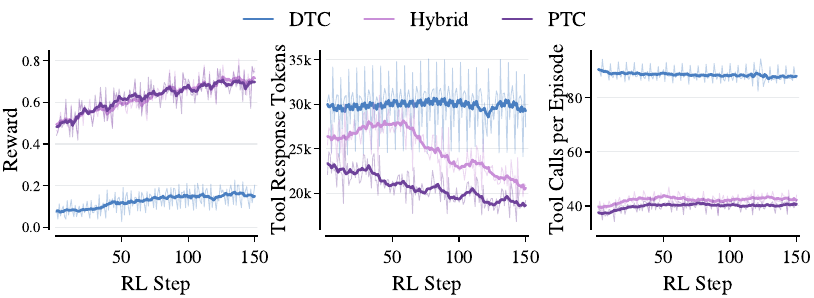}
\caption{\textbf{RL training dynamics at 14B.}
Episode reward (left), tool-output tokens entering context (middle), and tool calls per episode (right) under DTC, Hybrid, and PTC.
}
\label{fig:training_dynamics}
\end{figure}
\subsection{What Does PTC Change During Training?}
\label{sec:training_analysis}
To understand how PTC changes agent behavior, we track the RL training dynamics of Qwen3-14B in Figure~\ref{fig:training_dynamics}.
DTC achieves substantially lower rewards than PTC and Hybrid on our environments.
Specifically, DTC's reward increases only from 0.08 to 0.15, whereas the rewards of PTC and Hybrid start near 0.50 and reach 0.70 and 0.72, respectively.
For DTC, raw tool outputs account for around 30k context tokens per episode throughout training.
Programmatic agents instead express orchestration as executable code, retain intermediate state in sandbox variables, and selectively print results.
During RL training, the number of tool-output tokens under PTC decreases by roughly 20\%, falling below 20k per episode; both programmatic paradigms also use roughly half as many tool calls as DTC.
PTC runtime errors per episode also decrease from 6.4 to 3.6 at 14B and from 7.5 to 5.0 at 8B.
The 8B backbone reproduces the same trend (Appendix~\ref{sec:appendix_training_dynamics_8b}).

\subsection{Does Environment State Scaling Help?}
\begin{table}[h]
\centering
\caption{\textbf{Environment state and task difficulty.}
We present database size (KB), mean workspace file count, and DeepSeek V4 Pro's DTC pass@8 across the four environment scales.
}
\label{tab:env_scaling_stats}
\begin{tabular}{lccc}
\toprule
\textbf{Scale} & \textbf{Database Size (KB)} & \textbf{Workspace Files} & \textbf{pass@8} \\
\midrule
$v_0$ & 99  & 12.7 & 0.95 \\
$v_1$ & 173 & 21.9 & 0.88 \\
$v_2$ & 335 & 31.1 & 0.65 \\
$v_3$ & 711 & 40.3 & 0.42 \\
\bottomrule
\end{tabular}
\end{table}

Environment state scaling provides a direct mechanism to control task difficulty for the DTC harness.
As the environment state scales from $v_0$ to $v_3$, DeepSeek V4 Pro's pass@8 under the DTC harness drops monotonically from 0.95 to 0.42 in Table~\ref{tab:env_scaling_stats}.
As shown in Table~\ref{tab:ablations_loca}, restricting training from all four scales to the initial environments $v_0$ reduces average pass rates on LOCA-bench from 24.3\% to 16.0\% at 8B and from 38.3\% to 23.7\% at 14B.
These results show that environment state scaling improves PTC performance on LOCA-bench across both backbones.

\subsection{How Does PTC Facilitate Tool Orchestration?}
\label{sec:what_transfers}

\begin{table}[ht]
\centering
\caption{\textbf{Programming structures used during RL.}
We present the percentage of generated code blocks using loops and conditional execution, and the percentage of executions using error handling.}
\label{tab:programming_structures}
\small
\setlength{\tabcolsep}{5pt}
\begin{tabular}{lccc}
\toprule
\textbf{Model} & \textbf{Loops} & \textbf{Conditional execution} & \textbf{Error handling} \\
\midrule
8B & 82.6 & 65.6 & 8.2 \\
14B & 82.1 & 64.0 & 9.5 \\
\bottomrule
\end{tabular}
\end{table}

As shown in Table~\ref{tab:programming_structures}, PTC models use a range of programming structures, including loops, conditional execution, and error handling, to orchestrate tools during RL.
The Toolathlon-Verified example (Appendix~\ref{sec:case_toolathlon}) shows how PTC uses these programming structures to coordinate dependent tool calls within a program.
The MCP-Atlas example (Appendix~\ref{sec:case_atlas}) illustrates another use of PTC: aggregating intermediate tool results and selectively printing a summary into the model context.

\begin{table}[!t]
  \centering
  \caption{
      \textbf{Cross-harness evaluation on LOCA-bench and MCP-Atlas.}
      Pass rates (\%) under DTC and PTC evaluation for untrained backbones (Base) and models trained with the DTC or PTC SFT+RL recipe.
      LOCA-bench scores are averaged over four environment-state sizes. 
  }
  \label{tab:dtc_transfer_loca}
  \renewcommand{\arraystretch}{1.1}
  \begin{tabular}{@{}ll ccc ccc@{}}
  \toprule
  \multirow{2}[2]{*}{\textbf{Model}} & \multirow{2}[2]{*}{\textbf{Eval}} &
  \multicolumn{3}{c}{\textbf{LOCA-bench}} & \multicolumn{3}{c}{\textbf{MCP-Atlas}} \\
  \cmidrule(lr){3-5} \cmidrule(lr){6-8}
   & & Base & DTC-trained & PTC-trained & Base & DTC-trained & PTC-trained \\
  \midrule
  8B & DTC & 0.3 & 4.0 & 6.0 & 7.6 & 19.0 & 30.6 \\
   & PTC & 0.0 & 10.7 & 24.3 & 3.6 & 11.8 & 38.6 \\
  \midrule
  14B & DTC & 0.3 & 9.0 & 13.9 & 12.4 & 29.7 & 41.6 \\
   & PTC & 0.0 & 20.7 & 38.3 & 3.8 & 29.0 & 45.6 \\
  \bottomrule
  \end{tabular}
\end{table}

We next examine whether the benefits of PTC training extend beyond the programmatic interface.
To separate training effects from harness effects, we evaluate DTC- and PTC-trained models under both DTC and PTC harnesses on LOCA-bench and MCP-Atlas (Table~\ref{tab:dtc_transfer_loca}).
For DTC-trained models, switching from the DTC to the PTC harness improves performance on LOCA-bench but not on MCP-Atlas.
We also evaluate other baselines under the PTC harness.
Their pass rates remain low (Table~\ref{tab:ablations_loca}), suggesting that access to the PTC interface alone is insufficient for effective orchestration. 

Under the same PTC harness, PTC-trained models outperform DTC-trained models on both benchmarks.
This advantage also holds under DTC evaluation: on MCP-Atlas, the 14B PTC-trained model outperforms its DTC-trained counterpart by 11.9 percentage points.
These results suggest that PTC training improves tool orchestration beyond the gains from the PTC harness alone, with benefits that extend to direct tool calling.
This advantage persists under $v_0$-only training (Table~\ref{tab:ablations_loca}).

\subsection{What Challenges Remain for PTC Agents?}
\label{sec:when_ptc_helps}

Despite the overall gains in \S\ref{sec:main_results}, our analysis identifies two limitations of the current PTC agents.

\textbf{DTC-like behavior in complex environments}: When environments return unpredictable or highly unstructured content, PTC agents may revert to DTC-like behavior.
On Playwright, choosing the next tool call may require interpreting the current snapshot.
For \method{}-14B, 40.7\% of PTC calls contain a single tool invocation without control flow, error handling, or selective printing (Table~\ref{tab:mcpmark_interaction_costs}).
This behavior limits the benefits of programmatic orchestration and context management.

\textbf{Silent server failures}: The PTC harness of \method{} relies on MCP's \texttt{isError} field to raise tool exceptions;
however, non-compliant servers often omit this field, bypassing the agent's exception-based recovery.
Disabling this server-level error detection drops LOCA-bench pass rates from 24.3\% to 19.3\% at 8B and from 38.3\% to 34.5\% at 14B (Table~\ref{tab:ablations_loca}).
Without explicit exceptions, the agent may process error messages as successful tool returns.

\section{Conclusion}
\label{sec:conclusion}

We introduce \method{}, a training stack that combines verifiable tasks at multiple environment scales, a lightweight PTC runtime, and a training recipe comprising cold-start SFT followed by RL.
Using 215 tasks at four scales, \method{} improves agentic performance across four benchmarks and outperforms the evaluated baselines of comparable size.
Our analysis shows that PTC agents selectively print outputs and use programming structures to coordinate tool calls.
Cross-harness evaluations show that the benefits of PTC training extend to DTC. 

These findings suggest that PTC is useful not only as an inference interface but also as a training paradigm for tool orchestration.
The transfer of these gains to DTC motivates further study of how programmatic training shapes tool-use behavior across interfaces.
Our task synthesis pipeline and runtime provide a basis for studying this question across environment scales and training settings.
We hope \method{} will have a positive and lasting impact on long-horizon agent research.

\section*{AI Use Statement}
\label{sec:ai_use_statement}
We used generative AI tools for language editing, synthetic task construction and auditing, and trajectory generation for filtering and training.
The authors reviewed all AI-assisted content and take full responsibility for this work.

\bibliography{iclr2027_conference}
\bibliographystyle{iclr2027_conference}

\clearpage
\appendix
\section{Related Work}
\label{sec:related}

\paragraph{Programmatic tool calling.}
Executable code allows an agent to compose multiple tool calls within a single action~\citep{patel2026bitterlesson}. 
CodeAct~\citep{wang2024executable} uses executable code as the agent's action format and demonstrates improvements over JSON-based tool calls. However, it does not provide verifiable long-horizon tasks, which are crucial for agentic training.
Recursive Language Models~\citep{zhang2025recursive} primarily target long-context processing, storing long inputs in REPL variables and processing selected snippets through code and recursive model calls.
In MCP settings, code-execution patterns expose external tools as callable APIs and process intermediate results within a code block to reduce token use and latency~\citep{anthropic_code_execution_mcp_2025, anthropic_advanced_tool_use_2025}.
\method{} studies PTC as a training paradigm for long-horizon tool orchestration.
We combine verifiable tasks at controlled environment scales with a runtime that isolates episodes and preserves variables across turns, supporting both SFT and RL.

\paragraph{Scalable Environment Synthesis.}
Recent efforts to scale training environments span simulated and executable settings.
Simulated environments use an LLM as a world model to generate tool returns, as in WebWorld~\citep{xiao2026webworld}, SWE-World~\citep{sun2026swe}, DreamGym~\citep{chen2026scaling}, and Qwen-AgentWorld~\citep{zuo2026qwen}.
Executable environments implement tool behavior through programs, database backends, or finite-state machines, using graph-based task synthesis~\citep{tu2026scaleenv, xu2026envfactory} or programmatic task synthesis~\citep{song2026envscaler, wang2026awm}.
Agent-World~\citep{dong2026agent} combines simulated and executable environments within a self-evolving training arena.
\method{} builds verifiable executable environments and controls task difficulty by scaling the environment state while keeping the instruction, tool interface, and verifier fixed across variants (\S\ref{sec:construction_pipeline}).
This design provides tasks at multiple scales for studying and training long-horizon tool orchestration.

\section{Details on Task Synthesis}
\label{sec:appendix_task_synthesis}

\subsection{Task and Tool Distributions}
\label{sec:appendix_intents}

\begin{figure}[h]
\centering
\includegraphics[width=\linewidth]{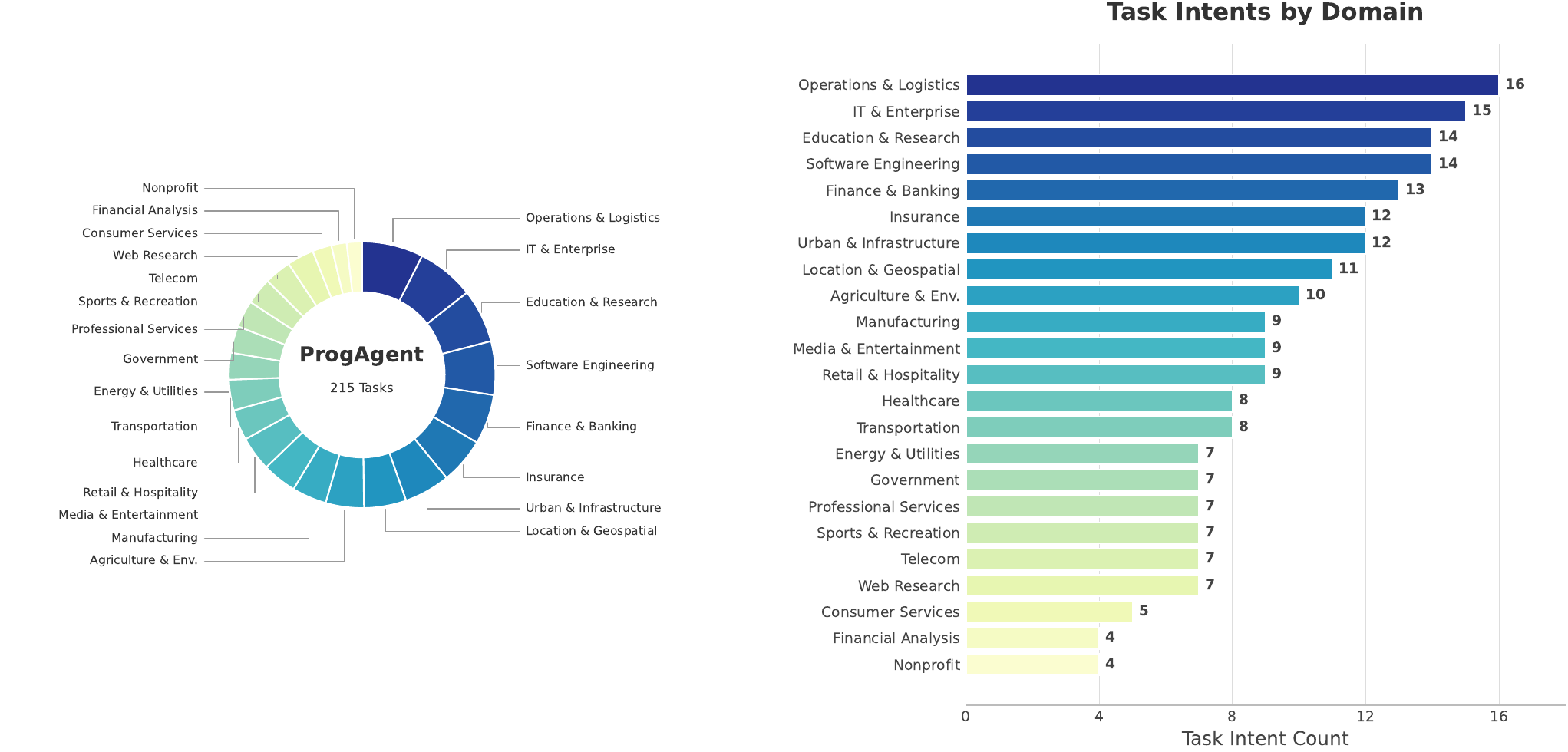}
\caption{Distribution of 215 tasks across 23 domains.}
\label{fig:topic_distribution}
\end{figure}

Figure~\ref{fig:topic_distribution} shows the distribution of the 215 tasks in \S\ref{sec:training} across 23 top-level domains.

\begin{figure}[h]
\centering
\includegraphics[width=0.8\linewidth]{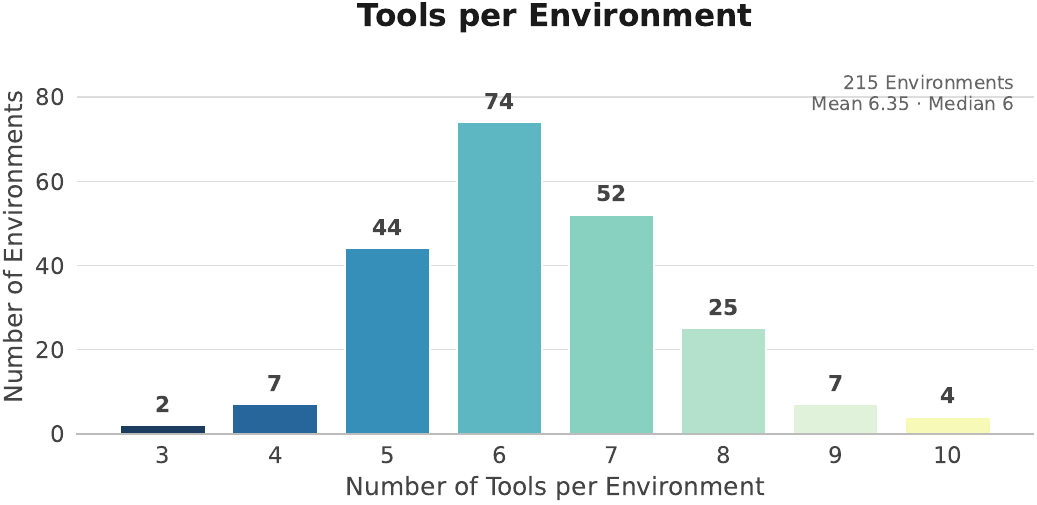}
\caption{Distribution of environment tool counts across the 215 tasks, excluding external harness tools. Each task is counted once.}
\label{fig:tool_count_distribution}
\end{figure}

Figure~\ref{fig:tool_count_distribution} shows the distribution of the number of environment tools available to the agent across the 215 tasks.
For each task, the tool set remains unchanged across all four environment scales.

\subsection{Task Construction Pipeline in Detail}
\label{sec:appendix_pipeline_details}
\begin{figure}[t]
\centering
\includegraphics[width=0.85\textwidth]{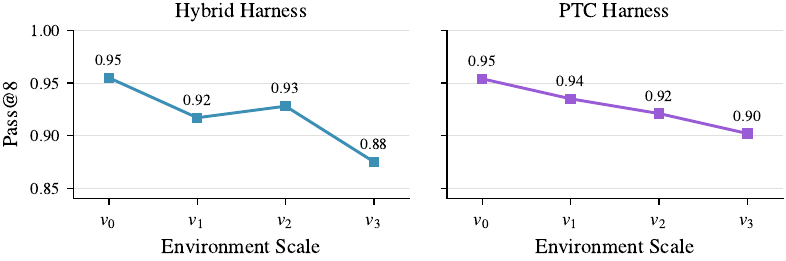}
\caption{\textbf{Environment scaling under Hybrid and PTC.}
DeepSeek V4 Pro's pass@8 on synthesized tasks at scales $v_0$--$v_3$; the DTC counterpart appears in Figure~\ref{fig:task_synthesis_pipeline}.}
\label{fig:env_scaling_harness}
\end{figure}

We describe the five task construction stages summarized in \S\ref{sec:construction_pipeline} in detail.

\textbf{Stage I: Domain collection and intent generation.}
We curate 23 top-level domains and their subcategories.
For each subcategory, an LLM generates user intents (prompt in \S\ref{sec:appendix_synthesis_prompts}).
A separate LLM call filters these candidates, retaining intents that involve sufficiently rich environment states for task construction.

\textbf{Stage II: Environment state construction.}
The constructor agent builds the initial environment state $s_0^E = (D, W)$, comprising a relational database $D$ and a file workspace $W$.
Solving each task requires combining records from $D$ with task constraints specified in unstructured files in $W$.
The constructor agent generates Python scripts to populate $D$ and $W$, with environment state size controlled by the \texttt{SCALE} configuration (\S\ref{sec:appendix_synthesis_prompts}).

\textbf{Stage III: Tool \& instruction synthesis.}
The constructor synthesizes the action space $\mathcal{A}_{\mathrm{tool}}$ as a set of single-purpose tools with exposed interface schemas and hidden Python implementations.
Each tool call returns a bounded amount of data, for example through pagination, so retrieving large amounts of data requires multiple calls.
The constructor also generates the task instruction $u_i$, specifying the task goals, constraints, and required outputs.
The instruction omits tool names, database schemas, and verifier implementation details; tool schemas are exposed separately through the harness.

\textbf{Stage IV: Verifier construction and task certification.}
Given the initial environment state $s_0^E$, $\mathcal{A}_{\mathrm{tool}}$, and $u_i$, the constructor instantiates a Python verifier $R_i(s_T^H, s_T^E) \in \{0,1\}$ that checks the final response and final environment state against the task's evaluation criteria. The verifier returns 1 if and only if all criteria are satisfied.
Verification does not use an LLM judge. The criterion-level score used for SFT trajectory selection is described in \S\ref{sec:appendix_synthesis_prompts}.
The constructor then generates a Python oracle solution restricted to the provided tools $\mathcal{A}_{\mathrm{tool}}$.
If the oracle fails verification, the constructor uses its execution trace to iteratively repair the environment data, tools, instruction, or verifier until the oracle receives a reward of 1.

\textbf{Stage V: Controlled scaling.}
The constructor varies the \texttt{SCALE} configuration from Stage~II to instantiate each task at four environment scales ($v_0$--$v_3$).
This expands the database and file workspace while keeping the task instruction, underlying logic, tool schemas, and verifier fixed.
The \texttt{SCALE} configuration sets database row counts to $1\times$, $2\times$, $4\times$, and $8\times$ their $v_0$ values, and scalable workspace file counts to $1\times$, $2\times$, $3\times$, and $4\times$, respectively.

As the environment scales from $v_0$ to $v_3$, DeepSeek V4 Pro's pass@8 decreases from 0.95 to 0.42 under DTC (Figure~\ref{fig:task_synthesis_pipeline}) while remaining above 0.87 under the Hybrid and PTC harnesses (Figure~\ref{fig:env_scaling_harness}).

\subsection{Verification-based Filtering in Detail}
\label{sec:appendix_filtering_details}

After oracle-based certification, candidate tasks may still contain inconsistencies between the instruction and verifier or fail to meet the difficulty criterion.
We therefore apply an independent harness audit followed by empirical difficulty filtering, as outlined in \S\ref{sec:filtering}.

\textbf{Independent harness audit.}
An independent harness agent writes and runs its own solver for each candidate task at the largest scale ($v_3$).
It then checks whether the tools, environment state, and verifier $R_i$ are consistent with the instruction $u_i$.
Tasks with identified issues are repaired and re-audited; tasks are discarded only during the subsequent difficulty filtering.

\textbf{Empirical difficulty filtering.}
After the audit, we evaluate task difficulty using model rollouts.
For each audited task $i$ and each scale $j \in \{0,1,2,3\}$, we sample $K=8$ independent trajectories from DeepSeek V4 Pro under the PTC harness.
We compute the pass rate $p_{i,j}$ as the fraction of trajectories for which the binary verifier $R_i$ returns 1.
We retain a task only if $p_{i,j} \in (0,0.8)$ at every scale.
This criterion removes 35 of the 250 candidate tasks, leaving 215 tasks.

For each retained task, we package each scale variant $\mathcal{M}_i^{(j)}$ into three artifacts in our native format:
a \emph{data generator} (a parameterized Python script that populates the database $D$ and workspace $W$),
an \emph{environment module} (tool implementations, the verifier $R_i$, and their enclosing environment class), and
the \emph{task prompt} $u_i$.
The runtime (\S\ref{sec:runtime}, Appendix~\ref{sec:appendix_runtime}) consumes these three artifacts.
We additionally provide each task in a Harbor-compatible format~\citep{harbor_framework_2026} for use outside our training stack.

\subsection{Prompts for Task Synthesis and Auditing}
\label{sec:appendix_synthesis_prompts}

In Stage~I, we use Gemini~3.0~Flash to generate user intents for each curated subcategory.

\begin{promptbox}{Intent generation prompt (Stage I)}
\small
\setlist[itemize]{leftmargin=1.2em,nosep}

Here are some existing task examples for reference:\\
\texttt{\{seed\_text\}}

Generate \texttt{\{num\_topics\}} brand-new task descriptions for the domain ``\texttt{\{category\}}''.

Requirements:
\begin{itemize}
  \item Length: 20-50 words.
  \item Realistic and plausible in everyday business/operations.
  \item Involves more than 3 data entities with meaningful data volume.
  \item Describes an operational workflow that concludes with an actionable verifiable output.
\end{itemize}

Output a JSON array only. Each element must have:
\begin{itemize}
  \item \texttt{"topic"}: the task topic description.
  \item \texttt{"folder"}: a unique short snake\_case directory name.
\end{itemize}

Output only the JSON array.
\end{promptbox}

For each retained intent, the constructor agent, powered by Claude Opus 4.6 through Claude Code, receives a single prompt covering Stages II--V of \S\ref{sec:construction_pipeline}.
The \texttt{step} method returns a criterion-level score $\widetilde R_i \in [0,1]$, and \texttt{verify} executes the oracle solution $\pi_i^\star$ and submits its output to \texttt{step}.
Difficulty filtering and RL use the binary verifier $R_i$, which returns 1 if and only if $\widetilde R_i=1$.
SFT trajectory selection uses the criterion-level score $\widetilde R_i$, retaining trajectories with $\widetilde R_i \geq 0.7$ (\S\ref{sec:appendix_hyperparams}).
We show excerpts from the task construction and harness audit prompts, retaining the opening instructions of each section and marking omitted details with ellipses.

\begin{promptbox}{Task construction prompt (Stages II--V, excerpts)}
\small

You can use the conda environment \texttt{eval} to build a complex Agent task. The task topic is:

\begin{center}\textbf{\texttt{\{TOPIC\}}}\end{center}

The Agent reads \texttt{prompt.md} as instruction and completes the task by calling tools defined in \texttt{env.py} and reading files in the \texttt{workspace}.

You are building the same task realized at multiple data scales (\texttt{v0} $\rightarrow$ \texttt{v3}); difficulty is introduced only by scale.

\promptsection{1. Build the Data Generation Script \texttt{data.py}}

Create \texttt{v0/data.py} to initialize the database and generate workspace files.

\ldots

\promptsection{2. Build the Environment \texttt{env.py}}

Implement \texttt{v0/env.py}. \texttt{env.py} and \texttt{prompt.md} will be copied byte-for-byte to v1/v2/v3; only \texttt{data.py}'s \texttt{SCALE} block differs across versions. A single implementation must therefore handle every scale from v0 to v3 without special-casing.

\ldots

\promptsection{3. Task Design}

\promptsubsection{3.1 Task Definition (\texttt{prompt.md})}

Write the task description \texttt{v0/prompt.md} based on the existing data and tools.

\ldots

\promptsubsection{3.2 Task Evaluation (\texttt{Task\_Env.step})}

Based on \texttt{prompt.md}, implement the \texttt{Task\_Env.step} function in \texttt{env.py}.

\ldots

\promptsubsection{3.3 Task Verification (\texttt{Task\_Env.verify})}

Implement the \texttt{verify} method in \texttt{Task\_Env} to prove end-to-end solvability.

\ldots

\promptsection{4. Replicate to v1, v2, v3}

Once \texttt{v0/} passes verification at the v3 scale, produce the remaining three versions:

\ldots

\end{promptbox}

An independent harness agent, powered by Claude Opus 4.6 through Claude Code, solves each task at $v_3$ using its own solver and checks the directory structure, instruction--verifier alignment, and consistency across scale variants.

\begin{promptbox}{Harness audit prompt (excerpts)}
\small
\setlist[enumerate]{leftmargin=1.5em,nosep}

You have just built a task at \texttt{v0/} and copied it to \texttt{v1/..v3/}. Your job is to verify the four versions hold together and fix any issues. You can use the conda environment \texttt{eval}.

Work through the sections below in order. If any check fails, fix it in \texttt{v0/} first (the single source of truth), re-propagate to \texttt{v1/..v3/}, then re-run the relevant check.

\promptsection{1. Structural Integrity}

Confirm without opening any Python files:

\ldots

\promptsection{2. Solvability Verification at v3 Scale}

Verify the task is solvable on the hardest version

\ldots

\promptsection{3. Env Quality}

Read \texttt{v3/prompt.md} and \texttt{v3/env.py}'s \texttt{step} function side by side.

\ldots

\promptsection{4. \texttt{prompt.md} Quality}

Formatting:
\begin{enumerate}
  \item No title or heading at the top of the file.
\end{enumerate}

\ldots

\promptsection{5. Final Clean-Up and Report}

\begin{enumerate}
  \item Remove runtime artifacts from every \texttt{v\_n/}: \texttt{data.db}, \texttt{workspace/}, \texttt{\_\_pycache\_\_/}.
\end{enumerate}

\ldots

\end{promptbox}

\section{Reproducibility and Implementation Details}
\label{sec:appendix_reproducibility}

\subsection{Agent Prompts and Action Formats}
\label{sec:appendix_agent_prompts}

During training, all three harnesses provide a terminal tool and a \texttt{claim\_done} tool.
Calling \texttt{claim\_done} ends the episode and submits the agent's final response for verification.

DTC and Hybrid share the following system prompt during training:

\begin{promptbox}{Agent system prompt (DTC and Hybrid)}
\small
You are a helpful assistant. Use the provided tools to complete the task. When you have finished the task, call the 'claim\_done' tool to submit your result.
\end{promptbox}

PTC uses the following system prompt during training:

\begin{promptbox}{Agent system prompt (PTC)}
\small
You are a helpful assistant. Complete the task using the provided tools.
Tools you may call directly:
\begin{itemize}[leftmargin=1.2em,nosep]
  \item \texttt{programmatic\_tool\_call} --- Python sandbox; the only way to invoke env tools.
  \item \texttt{terminal} --- shell in the workspace directory.
  \item \texttt{claim\_done} --- submit your final answer when the task is complete.
\end{itemize}
\end{promptbox}

DTC exposes task tools for direct calls, PTC requires calls through \texttt{programmatic\_tool\_call}, and Hybrid supports both.
The PTC tool description is shown below; the Hybrid version differs only in the opening paragraph.

\begin{promptbox}{Code-execution tool description (PTC)}
\small
\setlist[itemize]{leftmargin=1.2em,nosep}

Run Python that calls the tools listed above as \texttt{tools["tool\_name"](*args, **kwargs)}. \textbf{This is the ONLY way to invoke env tools} --- they cannot be called as standalone tool calls, so every env tool must go through this sandbox. State (variables, imports) persists across calls. Use \texttt{print()} to see output.

USE WHEN: loops, conditionals, error handling, or chaining multiple tool calls with intermediate processing.

Notes:
\begin{itemize}
  \item Code runs in the workspace directory and file writes are restricted to it, don't write to \texttt{/tmp}; always use absolute paths for file writes; os, json, csv, sys are pre-imported.
  \item Tools return native Python values; the type and structure vary by tool (e.g. dict, list, or str). Always print and inspect the first result before processing many items; do not assume a result is a list and loop over it.
  \item Very large printed output is truncated; print summaries rather than large raw data.
  \item Tools may raise an exception; wrap calls in try/except to handle failures.
\end{itemize}

Usage examples:

Batch + conditional workflow:
\begin{lstlisting}[language=Python]
print(type(tools["get_info"](id='A001')))  # inspect return type first, then loop
results = []
for item in ['A001', 'A002', 'A003']:
    info = tools["get_info"](id=item)
    if info.get('status') == 'active': # info is a dict, confirmed above
        results.append(tools["get_details"](id=item))
    else:
        print(f'Skipping {item}')
    print(f'Processed {item}')
print('Collected', len(results))
\end{lstlisting}

Error handling:
\begin{lstlisting}[language=Python]
ok, failed = [], []
for item_id in ['A001', 'A002', 'A003']:
    try:
        r = tools["get_info"](id=item_id)
        ok.append(r)
    except Exception as e:
        failed.append((item_id, str(e)))
    print(f'{len(ok)} ok, {len(failed)} failed')
print('Failed:', failed[:3] if failed else 'none')
\end{lstlisting}
\end{promptbox}

\begin{promptbox}{Code-execution tool description (Hybrid): opening paragraph}
\small
Run Python that calls the tools listed above as \texttt{tools["tool\_name"](*args, **kwargs)}. State (variables, imports) persists across calls. Use \texttt{print()} to see output.
\end{promptbox}

\subsection{PTC Runtime Architecture}
\label{sec:appendix_runtime}

\paragraph{Kernel and tool interface.}
Each episode uses a persistent IPython kernel that preserves variables across turns.
Tools are invoked as \texttt{tools["tool\_name"](*args, **kwargs)}.
We use dictionary syntax because MCP tool names may contain characters that are invalid in Python identifiers.
The kernel executes each program $\rho_t$ and returns its \texttt{stdout} as the observation $r_{\rho_t}$, or a traceback if execution fails.
Output is capped at 50,000 characters; longer outputs retain the first 35,000 and last 15,000 characters.

\paragraph{Sandbox layout.}
The kernel runs under Bubblewrap~\citep{bubblewrap2025} in separate user, mount, and PID namespaces.
The workspace $W$ is the only writable host-mounted path; system directories are read-only, and temporary storage is private to the sandbox.
The environment module and data generator are not mounted, preventing direct access to the verifier implementation and oracle solution.
Terminal commands use the same sandbox configuration.

\paragraph{Episode isolation and resource limits.}
Each episode starts with a fresh environment and kernel.
The task database is accessible only through environment tools.
Each kernel and its descendant processes are collectively limited to 2\,GiB of memory and 256 processes through cgroup~v2, with swap disabled.

\subsection{Training Hyperparameters}
\label{sec:appendix_hyperparams}

Both backbones, Qwen3-8B and Qwen3-14B, are used in thinking mode.
All three settings share the same pool of 215 tasks, each instantiated at four scales.
For cold-start SFT, we collect teacher trajectories separately under each harness and retain those with dense reward $\widetilde R_i \geq 0.7$ to obtain more training trajectories at a lower collection cost.
The three SFT datasets contain equal numbers of trajectories and each cover all 215 tasks.
We use the 0813 version of DeepSeek V4 Pro for all experiments.

\paragraph{Supervised fine-tuning.}
We apply YaRN~\citep{peng2024yarn} to extend the base model's native 32k context window to 128k for fine-tuning.
During SFT, we convert tool calls from JSON to XML format~\citep{qwen35blog} to avoid JSON escaping.
Models are trained with ms-swift~\citep{zhao2024swiftascalablelightweightinfrastructure} for 3 epochs at a batch size of 64, a learning rate of $1 \times 10^{-5}$, and a maximum sequence length of 128k.
SFT uses tensor parallelism of 2 with sequence parallelism on 2 nodes, each with 8 H200 GPUs.

\paragraph{Reinforcement learning.}
All three settings use GSPO~\citep{zheng2025stabilizing,gspo} with the same task pool and verifiers.
We set $\epsilon_{\text{low}} = 3 \times 10^{-4}$ and $\epsilon_{\text{high}} = 4 \times 10^{-4}$, use no KL penalty, and set the learning rate to $1 \times 10^{-6}$.
We train for 150 steps, sampling 16 rollouts per task for 32 tasks at each step, at temperature 1.0. For DTC RL, we only use $v_0$ environments, as the other environments are too difficult for the model to obtain any rewards.

The maximum prompt and response lengths are 4k and 85,000 tokens, respectively.
Training runs on verl~\citep{sheng2025hybridflow,yu2026dapo} with a Megatron-LM~\citep{megatron-lm} backend and vLLM~\citep{kwon2023efficient} rollouts.
RL uses tensor parallelism of 4 and context parallelism of 2 on 8 nodes, each with 8 H200 GPUs.

\subsection{Evaluation Protocol}
\label{sec:appendix_eval_protocol}

\paragraph{Harnesses.}
Our evaluations retain each benchmark's official MCP servers, verifier, system prompt, and task-completion logic.
For the main results, we use the DTC harness for DTC-trained models and for baselines and untrained backbones evaluated in our own runs.
Hybrid additionally exposes \texttt{programmatic\_tool\_call}, while PTC exposes the benchmark tool schemas, but requires calls through \texttt{programmatic\_tool\_call}.
During  evaluation, \texttt{programmatic\_tool\_call} has a 120-second timeout.
For the main results, each trained model is evaluated under its training harness.
Table~\ref{tab:ablations_loca} additionally reports cross-harness evaluations of our trained models, untrained backbones under all three harnesses, and selected trained baselines under PTC.
Following Agent-World~\citep{dong2026agent}, we use YaRN~\citep{peng2024yarn} to extend the native 32k context windows of open-source models to 128k.

\paragraph{Benchmark subsets.}
We exclude the MCPMark GitHub suite because repeated account suspensions during evaluation prevented consistent comparisons across models.
The reported MCPMark average is the mean over the Filesystem, Notion, Playwright, and PostgreSQL suites.
LOCA-bench is run at its four environment-state sizes (\S\ref{sec:experiments}).
For MCP-Atlas, we use the full task set from the final release before version 2.0, which is compatible with our Python-based evaluation harness.

\paragraph{Sources of reported scores.}
The MCPMark scores of the frontier proprietary models, the open-source foundation models, and the trained baselines other than EnvFactory-8B are taken from Agent-World~\citep{dong2026agent}.
For the frontier proprietary models, the LOCA-bench and MCP-Atlas scores are taken from the respective benchmark papers~\citep{zeng2026loca, bandi2026mcpatlas}.
In the Toolathlon column of Table~\ref{tab:main_results}, scores for GPT-5.2, Gemini-3 Pro, Claude Sonnet-4.5, and DeepSeek-V3.2 are from the original Toolathlon benchmark and marked with $\dagger$.
All other scores in that column are from Toolathlon-Verified.
All remaining scores are from our own evaluations.

\section{Extended Experimental Results}
\label{sec:appendix_extended_results}

\subsection{Full Ablation Results}
\label{sec:appendix_ablations}

Table~\ref{tab:ablations_loca} reports the full LOCA-bench ablation results, including the results summarized in Figure~\ref{fig:paradigm_ablation}.

\paragraph{Untrained base models.}
We evaluate the untrained 8B and 14B backbones under DTC, Hybrid, and PTC harnesses to establish performance before training.
Pass rates remain near zero under all three harnesses.

\paragraph{Trained baselines under the PTC harness.}
We evaluate selected AWM~\citep{wang2026awm}, EnvFactory~\citep{xu2026envfactory}, and EnvScaler~\citep{song2026envscaler} checkpoints under the PTC harness.
Their pass rates remain near zero, suggesting that access to the PTC interface alone is insufficient for effective tool orchestration, consistent with \S\ref{sec:what_transfers}.

\paragraph{Training paradigms, scaled environments ($v_0$--$v_3$).}
We report SFT and subsequent RL results for DTC, Hybrid, and PTC models trained on all four environment scales.
Cross-harness evaluations help separate training effects from harness effects; additional rows evaluate DTC- and PTC-trained models under the DTC harness with the LOCA-bench memory tool.

\paragraph{Initial environments ($v_0$) only.}
To control for training environment difficulty, we restrict SFT and RL for DTC and PTC to the $v_0$ versions of all tasks.
Both training paradigms cover all 215 tasks in SFT and RL, with equal numbers of SFT trajectories.
We evaluate the resulting checkpoints under each harness.
At 8B and 14B, PTC-trained models have higher average pass rates than DTC-trained models under the same harness after SFT and after RL.

\paragraph{Server-level error detection disabled.}
We evaluate the final PTC checkpoints with the mechanism that converts MCP server errors into tool exceptions disabled (\S\ref{sec:when_ptc_helps}).
Average pass rates decrease at both model sizes, consistent with the importance of explicit exceptions for error recovery.

\begin{table}[h]
  \centering
  \caption{
      \textbf{Ablations on LOCA-bench.}
      Pass rates (\%) across four environment-state sizes.
      SFT and +RL denote checkpoints after supervised fine-tuning and subsequent reinforcement learning.
      For our trained models, Setting denotes the training paradigm; evaluation uses the same harness unless otherwise specified. Rows labeled ``memory tool'' use the DTC harness with the LOCA-bench memory tool.
      Values spanning the SFT and +RL columns report a single evaluated checkpoint.
  }
  \label{tab:ablations_loca}
  \setlength\tabcolsep{4pt}
  \renewcommand{\arraystretch}{1.1}
  \resizebox{\textwidth}{!}{%
  \begin{tabular}{l l cc cc cc cc cc}
  \toprule
  \multirow{2}[2]{*}{\textbf{Model}} &
  \multirow{2}[2]{*}{\textbf{Setting}} &
  \multicolumn{2}{c}{\textbf{8k}} &
  \multicolumn{2}{c}{\textbf{16k}} &
  \multicolumn{2}{c}{\textbf{32k}} &
  \multicolumn{2}{c}{\textbf{64k}} &
  \multicolumn{2}{c}{\textbf{Avg.}} \\
  \cmidrule(lr){3-4} \cmidrule(lr){5-6} \cmidrule(lr){7-8} \cmidrule(lr){9-10} \cmidrule(lr){11-12}
   & & SFT & +RL & SFT & +RL & SFT & +RL & SFT & +RL & SFT & +RL \\
  \midrule
  \multicolumn{12}{l}{\textit{Untrained base models}} \\
  \midrule
  \multirow{3}{*}{8B} & DTC & \multicolumn{2}{c}{1.3} & \multicolumn{2}{c}{0.0} & \multicolumn{2}{c}{0.0} & \multicolumn{2}{c}{0.0} & \multicolumn{2}{c}{0.3} \\
   & Hybrid & \multicolumn{2}{c}{1.3} & \multicolumn{2}{c}{0.0} & \multicolumn{2}{c}{0.0} & \multicolumn{2}{c}{0.0} & \multicolumn{2}{c}{0.3} \\
   & PTC & \multicolumn{2}{c}{0.0} & \multicolumn{2}{c}{0.0} & \multicolumn{2}{c}{0.0} & \multicolumn{2}{c}{0.0} & \multicolumn{2}{c}{0.0} \\
  \cmidrule(l){2-12}
  \multirow{3}{*}{14B} & DTC & \multicolumn{2}{c}{1.3} & \multicolumn{2}{c}{0.0} & \multicolumn{2}{c}{0.0} & \multicolumn{2}{c}{0.0} & \multicolumn{2}{c}{0.3} \\
   & Hybrid & \multicolumn{2}{c}{2.7} & \multicolumn{2}{c}{0.0} & \multicolumn{2}{c}{0.0} & \multicolumn{2}{c}{0.0} & \multicolumn{2}{c}{0.7} \\
   & PTC & \multicolumn{2}{c}{0.0} & \multicolumn{2}{c}{0.0} & \multicolumn{2}{c}{0.0} & \multicolumn{2}{c}{0.0} & \multicolumn{2}{c}{0.0} \\
  \midrule
  \multicolumn{12}{l}{\textit{Trained baselines under the PTC harness}} \\
  \midrule
  AWM-8B & PTC harness & \multicolumn{2}{c}{0.0} & \multicolumn{2}{c}{0.0} & \multicolumn{2}{c}{0.0} & \multicolumn{2}{c}{0.0} & \multicolumn{2}{c}{0.0} \\
  AWM-14B & PTC harness & \multicolumn{2}{c}{0.0} & \multicolumn{2}{c}{0.0} & \multicolumn{2}{c}{0.0} & \multicolumn{2}{c}{0.0} & \multicolumn{2}{c}{0.0} \\
  EnvFactory-8B & PTC harness & \multicolumn{2}{c}{0.0} & \multicolumn{2}{c}{0.0} & \multicolumn{2}{c}{0.0} & \multicolumn{2}{c}{0.0} & \multicolumn{2}{c}{0.0} \\
  EnvScaler-8B & PTC harness & \multicolumn{2}{c}{1.3} & \multicolumn{2}{c}{0.0} & \multicolumn{2}{c}{0.0} & \multicolumn{2}{c}{0.0} & \multicolumn{2}{c}{0.3} \\
  \midrule
  \multicolumn{12}{l}{\textit{Training paradigms, scaled environments ($v_0$--$v_3$)}} \\
  \midrule
  \multirow{8}{*}{8B} & DTC & 5.3 & 5.3 & 6.7 & 6.7 & 0.0 & 1.3 & 0.0 & 2.7 & 3.0 & 4.0 \\
   & DTC, memory tool & 6.7 & 4.0 & 1.3 & 4.0 & 0.0 & 1.3 & 0.0 & 1.3 & 2.0 & 2.7 \\
   & DTC, PTC harness & 16.0 & 20.0 & 10.7 & 10.7 & 8.0 & 8.0 & 1.3 & 4.0 & 9.0 & 10.7 \\
   & Hybrid, DTC harness & 10.7 & 13.3 & 6.7 & 8.0 & 6.7 & 4.0 & 4.0 & 4.0 & 7.0 & 7.3 \\
   & Hybrid & 22.7 & 28.0 & 24.0 & 28.0 & 13.3 & 16.0 & 10.7 & 14.7 & 17.7 & 21.7 \\
   & PTC, DTC harness & 5.3 & 17.3 & 4.0 & 5.3 & 0.0 & 0.0 & 0.0 & 1.3 & 2.3 & 6.0 \\
   & PTC, memory tool & 10.7 & 16.0 & 4.0 & 13.3 & 0.0 & 6.7 & 0.0 & 1.3 & 3.7 & 9.3 \\
   & PTC & 24.0 & 34.7 & 20.0 & 24.0 & 20.0 & 21.3 & 20.0 & 17.3 & 21.0 & 24.3 \\
  \cmidrule(l){2-12}
  \multirow{8}{*}{14B} & DTC & 8.0 & 14.7 & 12.0 & 13.3 & 4.0 & 4.0 & 0.0 & 4.0 & 6.0 & 9.0 \\
   & DTC, memory tool & 16.0 & 20.0 & 10.7 & 13.3 & 2.7 & 4.0 & 4.0 & 4.0 & 8.3 & 10.3 \\
   & DTC, PTC harness & 25.3 & 28.0 & 24.0 & 24.0 & 10.7 & 17.3 & 13.3 & 13.3 & 18.3 & 20.7 \\
   & Hybrid, DTC harness & 10.7 & 26.7 & 10.7 & 12.0 & 6.7 & 13.3 & 4.0 & 2.7 & 8.0 & 13.7 \\
   & Hybrid & 26.7 & 38.7 & 25.3 & 30.7 & 14.7 & 33.3 & 12.0 & 22.7 & 19.7 & 31.3 \\
   & PTC, DTC harness & 12.0 & 23.7 & 5.3 & 18.7 & 4.0 & 8.0 & 2.7 & 5.3 & 6.0 & 13.9 \\
   & PTC, memory tool & 21.3 & 24.0 & 13.3 & 14.7 & 6.7 & 9.3 & 1.3 & 5.3 & 10.7 & 13.3 \\
   & PTC & 32.0 & 41.3 & 30.7 & 40.0 & 33.3 & 41.3 & 28.0 & 30.7 & 31.0 & 38.3 \\
  \midrule
  \multicolumn{12}{l}{\textit{Initial environments ($v_0$) only}} \\
  \midrule
  \multirow{4}{*}{8B} & DTC & 5.3 & 8.0 & 2.7 & 2.7 & 1.3 & 2.7 & 1.3 & 0.0 & 2.7 & 3.3 \\
   & DTC, PTC harness & 5.3 & 4.0 & 4.0 & 4.0 & 2.7 & 4.0 & 1.3 & 2.7 & 3.3 & 3.7 \\
   & PTC, DTC harness & 8.0 & 13.3 & 2.7 & 8.0 & 1.3 & 4.0 & 1.3 & 2.7 & 3.3 & 7.0 \\
   & PTC & 13.3 & 21.3 & 13.3 & 18.7 & 14.7 & 14.7 & 4.0 & 9.3 & 11.3 & 16.0 \\
  \cmidrule(l){2-12}
  \multirow{4}{*}{14B} & DTC & 8.0 & 13.3 & 10.7 & 8.0 & 4.0 & 4.0 & 0.0 & 1.3 & 5.7 & 6.7 \\
   & DTC, PTC harness & 16.0 & 20.0 & 17.3 & 20.0 & 6.7 & 14.7 & 1.3 & 10.7 & 10.3 & 16.3 \\
   & PTC, DTC harness & 18.7 & 20.0 & 10.7 & 13.3 & 5.3 & 6.7 & 2.7 & 2.7 & 9.3 & 10.7 \\
   & PTC & 24.0 & 25.3 & 26.7 & 26.7 & 21.3 & 26.7 & 13.3 & 16.0 & 21.3 & 23.7 \\
  \midrule
  \multicolumn{12}{l}{\textit{Server-level error detection disabled}} \\
  \midrule
  8B & PTC & \multicolumn{2}{c}{29.3} & \multicolumn{2}{c}{18.7} & \multicolumn{2}{c}{17.3} & \multicolumn{2}{c}{12.0} & \multicolumn{2}{c}{19.3} \\
  14B & PTC & \multicolumn{2}{c}{42.0} & \multicolumn{2}{c}{40.0} & \multicolumn{2}{c}{29.3} & \multicolumn{2}{c}{26.7} & \multicolumn{2}{c}{34.5} \\
  \bottomrule
  \end{tabular}
  }
\end{table}

\subsection{Interaction Costs on MCPMark}
\label{sec:appendix_mcpmark_interaction_costs}

\begin{table}
\centering
\caption{\textbf{Interaction costs of PTC on MCPMark.}
Tool calls and tool-output tokens are averaged per trajectory.
Context limit reports the fraction of trajectories reaching the context limit.
The ``DTC-like calls'' column reports the percentage of PTC calls containing a single explicit tool invocation without control flow, error handling, or selective printing; any printed tool return is reproduced in full without processing.}
\label{tab:mcpmark_interaction_costs}
\setlength{\tabcolsep}{4pt}
\begin{tabular}{@{}llrrrr@{}}
\toprule
\textbf{Model} & \textbf{Environment} & \shortstack{\textbf{Tool}\\\textbf{calls}} & \shortstack{\textbf{Tool-output}\\\textbf{tokens}} & \shortstack{\textbf{Context}\\\textbf{limit (\%)}} & \shortstack{\textbf{DTC-like }\\\textbf{calls (\%)}} \\
\midrule
\multirow{4}{*}{8B}
 & Filesystem & 50.5 & 21,086 & 16.7 & 6.21 \\
 & Postgres & 68.3 & 22,404 & 36.9 & 18.00 \\
 & Notion & 78.7 & 39,259 & 69.6 & 10.26 \\
 & Playwright & 122.1 & 37,551 & 64.0 & 34.71 \\
\midrule
\multirow{4}{*}{14B}
 & Filesystem & 34.9 & 15,925 & 8.8 & 10.10 \\
 & Postgres & 58.0 & 22,200 & 8.3 & 14.70 \\
 & Notion & 66.7 & 35,992 & 46.4 & 18.71 \\
 & Playwright & 135.0 & 35,988 & 52.0 & 40.72 \\
\bottomrule
\end{tabular}
\end{table}

Table~\ref{tab:mcpmark_interaction_costs} reports interaction costs for the 8B and 14B PTC models across four MCPMark subsets.
At both model sizes, Playwright has the highest proportion of DTC-like PTC calls (\S\ref{sec:when_ptc_helps}).
It also has more tool-output tokens per trajectory and a higher fraction of trajectories reaching the context limit than Filesystem and PostgreSQL.

\subsection{Training Dynamics at 8B}
\label{sec:appendix_training_dynamics_8b}

\begin{figure}
\centering
\includegraphics[width=\textwidth]{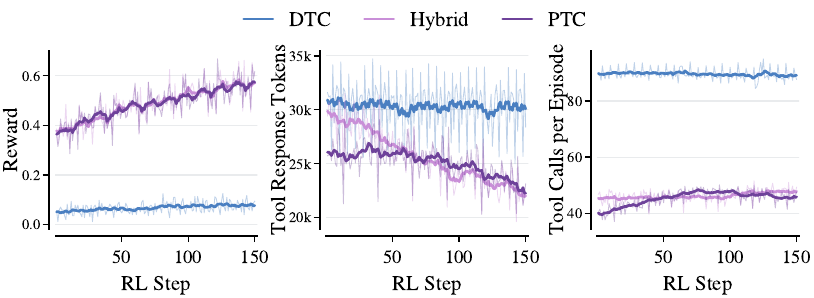}
\caption{\textbf{RL training dynamics at 8B.}
Episode reward (left), tool-output tokens entering context (middle), and tool calls per episode (right) under DTC, Hybrid, and PTC, with the same conventions as Figure~\ref{fig:training_dynamics}.}
\label{fig:training_dynamics_8b}
\end{figure}

Figure~\ref{fig:training_dynamics_8b} shows that the 8B backbone follows similar RL training trends to the 14B backbone (\S\ref{sec:training_analysis}).
Episode reward increases from approximately 0.05 to 0.08 under DTC, compared with 0.38 to 0.56 under both PTC and Hybrid.
Tool-output tokens entering context remain near 30k per episode under DTC, but decrease from 26.2k to 22.5k under PTC and from 29.6k to 22.1k under Hybrid.
PTC and Hybrid use roughly half as many tool calls per episode as DTC, approximately 45 versus 90.

\clearpage
\section{Qualitative Examples of PTC}
\label{sec:appendix_case_studies}

We present two concrete examples to illustrate how PTC works in practice.
For each task, we show the key steps of PTC execution, with \texttt{...} marking omitted output.

\begingroup
\definecolor{caseblue}{HTML}{315E96}
\definecolor{casebluebg}{HTML}{F3F6FB}
\definecolor{casegreen}{HTML}{35675A}
\definecolor{casegreenbg}{HTML}{F2F7F4}
\tcbset{casepanel/.style={enhanced, breakable, colback=white,
  colframe=promptframe, colbacktitle=prompttitlebg, coltitle=promptframe,
  fonttitle=\bfseries\small, fontupper=\small, boxrule=0.5pt, arc=2pt,
  left=7pt, right=7pt, top=5pt, bottom=5pt,
  before skip=7pt, after skip=7pt, parbox=false}}
\newcommand{\casetag}[4]{%
  \par\smallskip\noindent
  \tcbox[on line, colback=#2, colframe=#2, boxrule=0pt, arc=2pt,
    boxsep=0pt, left=5pt, right=5pt, top=2pt, bottom=2pt]{%
    \sffamily\bfseries\scriptsize\textcolor{#1}{#3}}%
  \hfill{\sffamily\scriptsize\textcolor{black!55}{#4}}\par\nobreak}
\lstset{basicstyle=\scriptsize\ttfamily\color{codeident},
  aboveskip=4pt, belowskip=4pt, lineskip=0pt}

\subsection{MCP-Atlas: From Campaign Records to a Statistical Answer}
\label{sec:case_atlas}

This case shows how PTC aggregates structured tool results inside the execution environment and returns a compact statistical summary.
Steps 1--3 are excerpts from a single PTC execution; its output is split between Steps 2 and 3 to match the code.

\begin{tcolorbox}[casepanel,title={Instruction}]
I'm trying to understand some ad data in my database. I want to see some details of campaigns that spend less than \$800, specifically what was the most common type of ad, and for the least used one, what the average audience reach was.
\tcblower
\casetag{caseblue}{casebluebg}{AVAILABLE MCP SERVERS}{}
{\sffamily\scriptsize Notion, Memory, MCP Code Executor.\par}
\end{tcolorbox}

\begin{tcolorbox}[casepanel,title={1. Query campaigns spending less than \$800}]
\casetag{caseblue}{casebluebg}{PTC CALL}{Python}
\begin{lstlisting}[language=Python]
# db_id: advertising database identified in earlier calls.
results = tools["notion_API-post-database-query"](
    database_id=db_id,
    filter={
        "property": "SpentAmount",
        "number": {"less_than": 800}
    }
)
\end{lstlisting}
\end{tcolorbox}

\begin{tcolorbox}[casepanel,title={2. Aggregate the records}]
\casetag{caseblue}{casebluebg}{PTC CALL}{Python}
\begin{lstlisting}[language=Python]
# Continuation of the same PTC call; results holds the query response.
from collections import Counter
ad_counts = Counter()
audience_reach_by_type = Counter()

for r in results['results']:
    ad_type = r['properties']['AdType']['select']['name']
    ad_counts[ad_type] += 1
    audience_reach_by_type[ad_type] += r['properties']['AudienceReach']['number']

print("Ad type counts:")
for ad_type, count in ad_counts.most_common():
    avg_reach = audience_reach_by_type[ad_type] / count
    print(f"  {ad_type}: {count} campaigns, avg audience reach={avg_reach:.1f}")
\end{lstlisting}
\casetag{casegreen}{casegreenbg}{OUTPUT}{Execution result}
\begin{lstlisting}[language={}]
Ad type counts:
  Social Media: 30 campaigns, avg audience reach=23512.0
  Search Engine: 28 campaigns, avg audience reach=24872.6
  Video: 20 campaigns, avg audience reach=25063.5
  Banner: 17 campaigns, avg audience reach=19380.5
\end{lstlisting}
\end{tcolorbox}

\begin{tcolorbox}[casepanel,title={3. Print the computed answer}]
\casetag{caseblue}{casebluebg}{PTC CALL}{Python}
\begin{lstlisting}[language=Python]
print("\nMost common type:", ad_counts.most_common(1)[0][0])
print("Least common type:", ad_counts.most_common()[-1][0])

least_common_type = ad_counts.most_common()[-1][0]
print(f"\nAverage audience reach for {least_common_type}: {audience_reach_by_type[least_common_type] / ad_counts[least_common_type]:.1f}")
\end{lstlisting}
\casetag{casegreen}{casegreenbg}{OUTPUT}{Execution result}
\begin{lstlisting}[language={}]
Most common type: Social Media
Least common type: Banner

Average audience reach for Banner: 19380.5
\end{lstlisting}
\end{tcolorbox}

\begin{tcolorbox}[casepanel,title={4. Check completeness before the final response}]
\casetag{caseblue}{casebluebg}{PTC CALL}{Python}
\begin{lstlisting}[language=Python]
results = tools["notion_API-post-database-query"](
    database_id=db_id,
    filter={
        "property": "SpentAmount",
        "number": {
            "less_than": 800
        }
    },
    page_size=100
)
print(f"Total: {len(results.get('results', []))}, Has more: {results.get('has_more', False)}")

# Also let's verify by looking at min and max spent
spent_values = []
ad_types = []
for r in results['results']:
    spent_values.append(r['properties']['SpentAmount']['number'])
    ad_types.append(r['properties']['AdType']['select']['name'])

print(f"Min spent: {min(spent_values)}, Max spent: {max(spent_values)}")
print(f"Ad types: {set(ad_types)}")
\end{lstlisting}
\casetag{casegreen}{casegreenbg}{OUTPUT}{Execution result}
\begin{lstlisting}[language={}]
Total: 95, Has more: False
Min spent: 509, Max spent: 798
Ad types: {'Banner', 'Social Media', 'Search Engine', 'Video'}
\end{lstlisting}
\end{tcolorbox}

\begin{tcolorbox}[casepanel,title={Final agent response}]
The analysis of campaigns spending less than \$800 is complete. Here's the summary:

\smallskip\hrule\smallskip
\textbf{Ad Campaign Analysis (Spent less than \$800)}

\begin{center}
\begin{tabular}{lrr}
\textbf{Ad Type} & \textbf{Campaigns} & \textbf{Avg Audience Reach} \\
\hline
Social Media & 30 & 23,512.0 \\
Search Engine & 28 & 24,872.6 \\
Video & 20 & 25,063.5 \\
\textbf{Banner} & \textbf{17} & \textbf{19,380.5} \\
\end{tabular}
\end{center}
\smallskip\hrule\smallskip
\textbf{Key Findings}
\begin{itemize}[leftmargin=1.3em,itemsep=2pt,topsep=3pt]
\item \textbf{Most common type of ad}: \textbf{Social Media} accounts for 30 of the 95 campaigns (31.6\%), followed by Search Engine (28) and Video (20).
\item \textbf{Least used type}: \textbf{Banner} ads appear in just 17 campaigns (17.9\%).
\item The \textbf{least used type --- Banner} --- has an average audience reach of \textbf{19,380.5}.
\end{itemize}
\end{tcolorbox}

\newpage
\subsection{Toolathlon-Verified: From an Emailed Schedule to Published Canvas Courses}
\label{sec:case_toolathlon}

This case shows how PTC uses loops and returned identifiers to coordinate dependent course-creation and configuration calls.

\begin{tcolorbox}[casepanel,title={Instruction}]
This is admin3, a registrar from XXX University. The university has sent me the new semester's course schedule via email with the subject "Course Schedule Notification." I need you to help me find the courses I'm responsible for from the attached master schedule and create corresponding courses on Canvas for the instructors. The specific requirements are as follows:
\begin{itemize}[leftmargin=1.3em,itemsep=2pt,topsep=4pt]
\item Assign the corresponding instructor as the teacher for each course on Canvas. The instructors' emails are saved in my local workspace.
\item Publish an announcement for each course that includes the instructor's name and the class time.
\item If an instructor is teaching the same course multiple times, these courses should be set up for resource sharing in Canvas.
\item The syllabi for each course are saved in my workspace. Please publish them for their respective courses.
\item All courses I am responsible for need to be published.
\end{itemize}
\tcblower
\casetag{caseblue}{casebluebg}{AVAILABLE MCP SERVERS}{}
{\sffamily\scriptsize Filesystem, Canvas, Terminal, Emails.\par}
\end{tcolorbox}

\begin{tcolorbox}[casepanel,title={1. Select assigned courses and look up instructor IDs}]
\casetag{caseblue}{casebluebg}{PTC CALL}{Python}
\begin{lstlisting}[language=Python]
# data: parsed schedule rows (course, instructor, time, administrator).
# parse_tool_result: helper defined earlier to decode the tool response.
admin3_courses = [(c, i, t) for c, i, t, a in data if a == "admin3"]
instructor_map = {}
for name in set(i for _, i, _ in admin3_courses):
    result = tools["canvas_canvas_list_account_users"](
        account_id=account_id, search_term=name, page=1, per_page=20)
    users = parse_tool_result(result)
    if users:
        u = users[0] if isinstance(users, list) else users
        instructor_map[name] = u['id']
\end{lstlisting}
\end{tcolorbox}

\begin{tcolorbox}[casepanel,title={2. Create courses and use returned IDs to enroll instructors}]
\casetag{caseblue}{casebluebg}{PTC CALL}{Python}
\begin{lstlisting}[language=Python]
print(f"Processing {len(admin3_courses)} courses...")
created_courses = []
for idx, (course_name, instructor, class_time) in enumerate(admin3_courses):
    try:
        course = tools["canvas_canvas_create_course"](
            account_id=account_id, name=course_name,
            start_at=(date.today()).isoformat(),
            end_at=(date.today() + timedelta(days=365)).isoformat(),
            time_zone="America/Denver")
        course_id = course['id']
        tools["canvas_canvas_enroll_user"](
            course_id=course_id, user_id=instructor_map[instructor],
            role="TeacherEnrollment", enrollment_state="active")
        created_courses.append((course_name, course_id, instructor, class_time))
        print(f"[{idx+1}/{len(admin3_courses)}] Created {course_name} (ID={course_id}) for {instructor}")
    except Exception as e:
        print(f"ERROR {idx}: {e}")
print(f"\nCreated {len(created_courses)} courses.")
\end{lstlisting}
\casetag{casegreen}{casegreenbg}{OUTPUT}{Execution result}
\begin{lstlisting}[language={}]
Processing 27 courses...
[1/27] Created English Literature-7 (ID=31) for Melissa Sanchez
[2/27] Created Business Management-7 (ID=32) for Cynthia Gomez
[3/27] Created Philosophy-7 (ID=33) for Donald Reed
[4/27] Created Creative Writing-7 (ID=34) for Carolyn Nguyen
...
Created 27 courses.
\end{lstlisting}
\end{tcolorbox}

\begin{tcolorbox}[casepanel,title={3. Create announcements using course IDs}]
\casetag{caseblue}{casebluebg}{PTC CALL}{Python}
\begin{lstlisting}[language=Python]
# created_courses: 27 course tuples listed earlier in this call.
for course_name, course_id, instructor, class_time in created_courses:
    message = f"Welcome to {course_name}. This course is taught by {instructor} at {class_time}."
    try:
        tools["canvas_canvas_create_announcement"](
            course_id=course_id, title=f"{course_name} Class Info",
            message=message, published=True)
        print(f"Announcement for {course_name} (ID={course_id}): {message[:80]}...")
    except Exception as e:
        print(f"Announcement ERROR for {course_name}: {e}")
\end{lstlisting}
\casetag{casegreen}{casegreenbg}{OUTPUT}{Execution result}
\begin{lstlisting}[language={}]
Announcement for English Literature-7 (ID=31): Welcome to English Literature-7. This course is taught by Melissa Sanchez at Mon...
Announcement for Business Management-7 (ID=32): Welcome to Business Management-7. This course is taught by Cynthia Gomez at Wedn...
Announcement for Philosophy-7 (ID=33): Welcome to Philosophy-7. This course is taught by Donald Reed at Friday 1,2,3....
...
\end{lstlisting}

\end{tcolorbox}

The following program checks course publication, teacher count, the public-syllabus flag, and announcement presence.
Its success message refers to these checks; it does not verify instructor identity, syllabus or announcement content, or resource sharing.

\begin{tcolorbox}[casepanel,title={4. Check selected course settings and summarize results}]
\casetag{caseblue}{casebluebg}{PTC CALL}{Python}
\begin{lstlisting}[language=Python,literate={✓}{{$\checkmark$}}1 {✗}{{$\times$}}1]
# courses: 27 course-name/ID pairs (literal list omitted).
# instructor_map: Canvas instructor IDs to names (dictionary omitted).
print("=== FINAL COMPREHENSIVE VERIFICATION ===\n")

# Track instructor-to-course count
instructor_courses = {}
for cname, cid in courses:
    detail = tools["canvas_canvas_get_course"](course_id=cid)
    tid = detail['teachers'][0]['id'] if detail['teachers'] else None
    if tid:
        instructor_courses.setdefault(tid, []).append(cname)

# Verify
all_ok = True
failures = []

for course_name, course_id in courses:
    detail = tools["canvas_canvas_get_course"](course_id=course_id)
    state_ok = detail['workflow_state'] == 'available'
    teacher_ok = len(detail['teachers']) == 1
    # Check teacher is correct
    actual_teacher = detail['teachers'][0]['display_name'] if detail['teachers'] else 'NONE'

    syll_ok = detail['public_syllabus']
    anns = tools["canvas_canvas_list_announcements"](course_id=course_id)
    ann_ok = len(anns) >= 1

    if not state_ok:
        failures.append(f"{course_name}: state={detail['workflow_state']}")
    if not teacher_ok:
        failures.append(f"{course_name}: teachers count={len(detail['teachers'])}")
    if not syll_ok:
        failures.append(f"{course_name}: syllabus not public")
    if not ann_ok:
        failures.append(f"{course_name}: no announcements")

    print(f"{course_name}: state={detail['workflow_state']}, teachers={len(detail['teachers'])} ({actual_teacher}), syll={syll_ok}, anns={len(anns)}")

if failures:
    print(f"\n✗ {len(failures)} failures: {failures}")
else:
    print(f"\n✓ All {len(courses)} courses verified successfully!")

# Instructor summary
print("\nInstructor-Course Mapping:")
for tid, cnames in sorted(instructor_courses.items()):
    print(f"  {tid} ({instructor_map[tid]}): {len(cnames)} courses - {cnames}")

print(f"\nTotal unique instructors: {len(instructor_courses)}")
print(f"Total courses: {len(courses)}")
\end{lstlisting}

\casetag{casegreen}{casegreenbg}{OUTPUT}{Execution result}
\begin{lstlisting}[language={},literate={✓}{{$\checkmark$}}1]
=== FINAL COMPREHENSIVE VERIFICATION ===

English Literature-7: state=available, teachers=1 (Melissa), syll=True, anns=1
Business Management-7: state=available, teachers=1 (Cynthia), syll=True, anns=1
Philosophy-7: state=available, teachers=1 (Donald), syll=True, anns=1
Creative Writing-7: state=available, teachers=1 (Carolyn), syll=True, anns=1
...

✓ All 27 courses verified successfully!

Instructor-Course Mapping:
  62 (Dennis Robinson): 3 courses - ['Marketing Principles-7', 'Operations Management-7', 'Innovation Management-7']
  63 (Donald Reed): 3 courses - ['Philosophy-7', 'Optics-7', 'Business Strategy-7']
...

Total unique instructors: 11
Total courses: 27
\end{lstlisting}
\end{tcolorbox}
\endgroup

\end{document}